\documentclass[lettersize,journal]{IEEEtran}
\usepackage{makecell}
\usepackage{paralist}
\usepackage{bbm}
\usepackage{xspace}             % 必须先加载，否则 \xspace 无效
\usepackage{amsmath,amsfonts}   % \eqref 来自 amsmath
\usepackage{algorithmic}
\usepackage{algorithm}
\usepackage{array}
\newcolumntype{I}{!{\vrule width 1pt}}
\usepackage[table,svgnames]{xcolor} % 解决 \rowcolor 和颜色名（如 RoyalBlue）
\usepackage{utfsym}                % 解决 \usym{2717} 这种特殊符号
\usepackage{cancel}                % 解决 \bcancel 画斜杠
\usepackage{makecell}              % 如果你用了 \Xhline，必须带上这个

\usepackage[caption=false,font=normalsize,labelfont=sf,textfont=sf]{subfig}
\usepackage{textcomp}
\usepackage{stfloats}
\usepackage{url}
\usepackage{verbatim}
\usepackage{graphicx}
\usepackage{cite}

\usepackage{xcolor}
\usepackage[dvipsnames]{xcolor}

\usepackage{overpic}     
\usepackage{multirow}   
\usepackage{multicol}    
\usepackage{colortbl}    
\usepackage{booktabs}    
\usepackage[colorlinks=true,linkcolor=blue,citecolor=blue,urlcolor=blue]{hyperref}
\usepackage[capitalize]{cleveref}
\usepackage{tikz,xcolor,hyperref}% Make Orcid icon
\definecolor{lime}{HTML}{A6CE39}
\DeclareRobustCommand{\orcidicon}{%
    \begin{tikzpicture}
    \draw[lime, fill=lime] (0,0) 
    circle [radius=0.16] 
    node[white] {{\fontfamily{qag}\selectfont \tiny ID}};    \draw[white, fill=white] (-0.0625,0.095) 
    circle [radius=0.007];    \end{tikzpicture}
    \hspace{-2mm}}
\foreach \x in {A, ..., Z}{%
    \expandafter\xdef\csname orcid\x\endcsname{\noexpand\href{https://orcid.org/\csname orcidauthor\x\endcsname}{\noexpand\orcidicon}}
    }
\makeatletter
\def\onedot{\ifx\@let@token.\else.\null\fi\xspace}
\makeatother

\definecolor{modifiedorange}{RGB}{255,140,0} % 橘⾊，可调整为你喜欢的⾊调
\definecolor{modifiedblue}{RGB}{64,120,192}
\definecolor{picred}{RGB}{153,34,46}
\begin{document}
% 放在导言区（\documentclass 后）
\setlength{\abovecaptionskip}{-5pt}  % caption 到“正文/图”的上方间距
\setlength{\belowcaptionskip}{0pt}  % caption 到下方内容的间距

\bstctlcite{BSTcontrol}

\title{DISTA-Net++: Rethinking Infrared Small Target Unmixing Beyond Sub-Pixel Separation
}

\author{
    Mengze Xu, 
    Zhu~Liu,
    Weidong Sheng,
    Boyang Li, 
    Yimian Dai,
    Ming-Ming Cheng,
    Jian Yang

\thanks{
This work was supported by the National Science Foundation of China (No. 62301261, % 我的青基
    No. 62225604, % 程老师杰青
    No. U24A20330 % 杨老师联合基金
    No. 62361166670), % 杨老师澳门基金
the Tianjin Natural Science Foundation Project (No. 25JCQNJC01370), % 我的天津市青基
the Shenzhen Science and Technology Program (No. JCYJ20240813114237048), % 程老师深圳市面上 
the Fundamental Research Funds for the Central Universities (No. 63261203, 
No. 63253217), % 程老师
and the Supercomputing Center of Nankai University (NKSC).
(Corresponding author: Yimian Dai).
An earlier version of this paper was presented at \href{https://openaccess.thecvf.com/content/ICCV2025/html/Han_DISTA-Net_Dynamic_Closely-Spaced_Infrared_Small_Target_Unmixing_ICCV_2025_paper.html}{ICCV 2025}. 

M. Xu, Y. Dai, M.-M. Cheng, and J. Yang are with VCIP, CS, Nankai University, Tianjin 300350, China (e-mail: mengze\_xu@mail.nankai.edu.cn; \{yimian.dai, cmm, csjyang\}@nankai.edu.cn).

Zhu Liu is with the School of Software Technology, Dalian University of Technology, Dalian 116024, China.  (e-mail:  liuzhucv@gmail.com).

W. Sheng and B. Li are with with the College of Electronic Science and Technology, National University of Defense Technology (NUDT), Changsha 410000, China (\{shengweidong, liboyang20\}@nudt.edu.cn).

}}

\maketitle

\begin{abstract}

Long-range infrared imaging frequently confronts dense target clusters whose diffraction-limited signatures merge into a single indistinguishable blob, concealing the number, sub-pixel positions, and radiant intensities of the underlying sources. 
While deep learning has advanced general object detection, resolving such Closely-Spaced Infrared Small Targets (CSIST) remains largely unexplored, owing to a systemic infrastructure void and a fundamental paradigm mismatch.
The dominant formulation, which reduces unmixing to a \textit{blind, discrete} sub-pixel separation, is inherently insufficient: without semantic guidance, the ill-posed inverse problem admits ambiguous solutions plagued by false and missed detections, while grid-based discretization locks predictions onto fixed lattice centers, chaining precision to prohibitively expensive grid refinement.
We argue that CSIST unmixing should instead be \textit{informed and continuous}. 
To ground this paradigm shift, we establish the first comprehensive open-source ecosystem for the field, comprising the large-scale CSIST-100K benchmark, a tailored metric suite, and the GrokCSO toolkit.
Upon this foundation, we propose DISTA-Net++, which anchors a dynamic deep unfolding backbone with two synergistic mechanisms: a Count-Guided Prior that injects the global target count as an explicit semantic constraint to regularize the solution space, and a Continuous Coordinate Rectification that regresses off-grid offsets to decouple localization accuracy from grid resolution.
Extensive experiments validate the proposed paradigm: even under the most economical 3× division, DISTA-Net++ surpasses 7×-division state-of-the-art methods by 16.15\% in CSO-mAP and 62.96\% in count accuracy at merely one-sixth of their computation, demonstrating that unmixing precision need not be purchased with finer discretization.
The complete ecosystem is released at \url{https://github.com/GrokCV/GrokDet}.

\end{abstract}


\section{Introduction}
\label{sec:intro}

\IEEEPARstart{I}{nfrared} imaging serves as a cornerstone of long-range surveillance and early-warning systems~\cite{TGRS2016TIRReview}, owing to its exceptional thermal sensitivity and its independence from illumination conditions. At operational distances, however, the radiation captured from targets is inherently weak, and targets degenerate into dim, point-like spots devoid of texture and structure~\cite{PR2023IRSTDSurvey}. Localizing such faint signatures against cluttered backgrounds is the central objective of Infrared Small Target Detection (IRSTD)~\cite{WACV2021ACM,zhang2025mirsam,li2026probing}, which constitutes the perceptual front-end of Infrared Search and Track (IRST) systems and underpins downstream modules such as multi-target tracking, trajectory prediction, and threat assessment.

\begin{figure}[t!]
\centering
\includegraphics[width=0.99\linewidth]{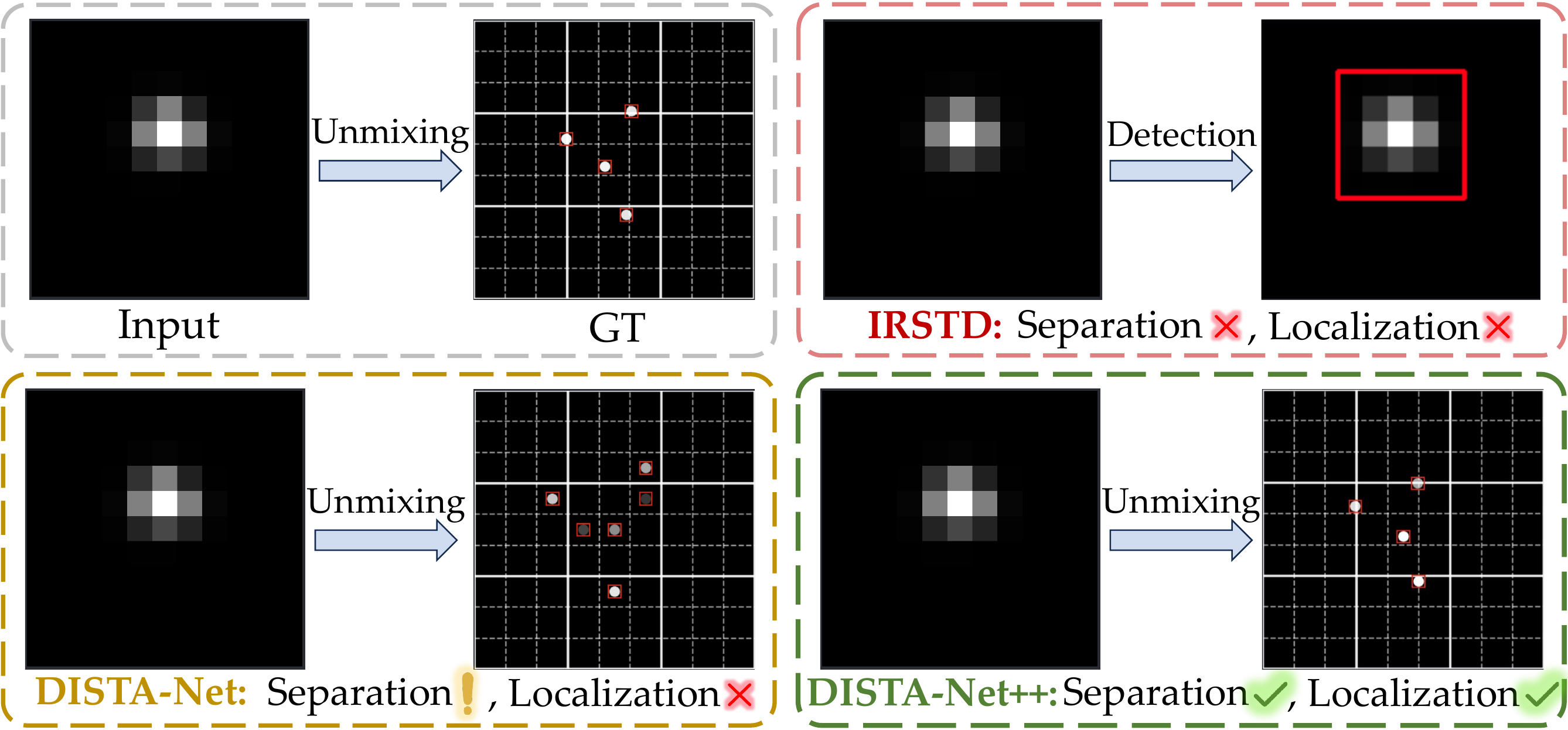}
\caption{
\textbf{Evolution of processing paradigms for closely-spaced infrared small 
targets (CSIST).} 
\textbf{Top right}: Infrared Small Target Detection (\textbf{IRSTD}) acts only as a coarse detector, reporting a single bounding box for overlapped targets. 
\textbf{Bottom left}: Existing unmixing models (e.g., \textbf{DISTA-Net}) succeed in 
separating the overlapped signals, yet in a \emph{blind, discrete} manner, 
thus suffering from count ambiguity and grid quantization error. 
\textbf{Bottom right:} The proposed \textbf{DISTA-Net++} establishes an \emph{informed, continuous} formulation, achieving faithful recovery of target count, sub-pixel positions, and intensities aligned with the Ground Truth (GT). Unmixing results are shown cropped and enlarged around the target region
for clarity.
}
\label{fig:pipeline_com}
\end{figure}

\begin{figure*}[!t]
    \centering
    \includegraphics[width=.92\textwidth]{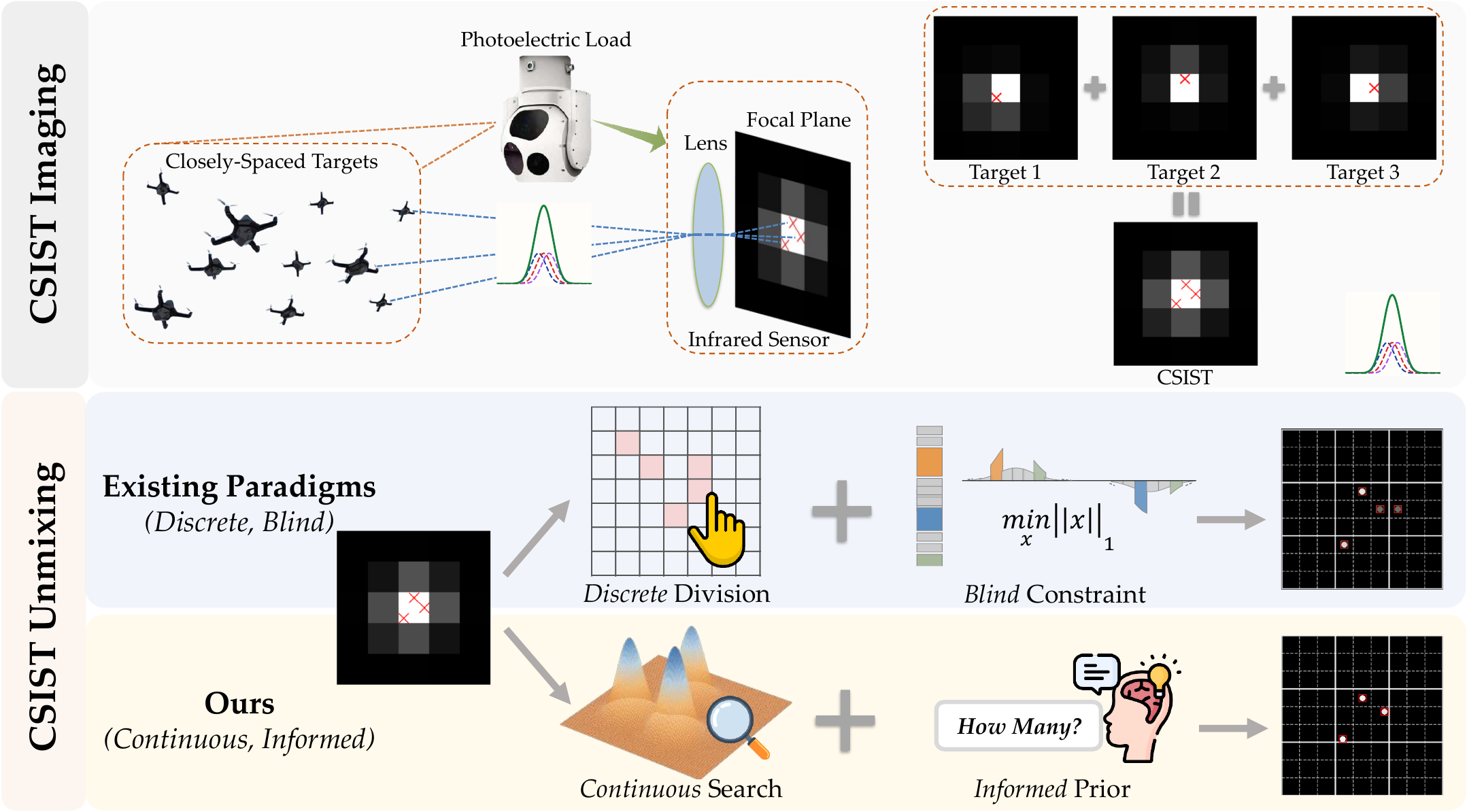}
    \caption{\textbf{Conceptual illustration of imaging and unmixing processes for closely-spaced infrared small targets (CSIST).} CSIST unmixing aims to disentangle and accurately estimate the count, positions, and intensities of overlapping targets.}
    \label{fig:CSIST-Imaging-Unmixing}
\end{figure*}

Yet a fundamental physical barrier lies beyond the reach of detection alone. 
When multiple targets approach one another within the diffraction limit of 
the optical system, their point spread functions inevitably overlap, 
collapsing distinct sources into a single, visually indivisible blob on the 
focal plane. Confronted with such Closely-Spaced Infrared Small Targets 
(CSIST)~\cite{AMOS2023MuyGPyS}, even a flawless IRSTD detector can do no 
better than report one coarse bounding box 
(Fig.~\ref{fig:pipeline_com}), while the number of underlying 
sources, their sub-pixel positions, and their radiant intensities all remain 
buried within the mixture~\cite{GRSM2022SingleFrameSurvey}. For an IRST 
system this failure is a silent one: a formation of threats is perceived as 
a single object, and no alarm is ever raised about the misperception. The 
task of \textit{CSIST unmixing}, namely recovering the count, locations, and 
intensities of individual sources from their superimposed observation 
(Fig.~\ref{fig:CSIST-Imaging-Unmixing}), is therefore not an incremental 
refinement of detection. It is a qualitatively distinct inverse problem that 
separates seeing a blob from understanding a cluster.

Research directly addressing this problem remains scarce, and virtually all existing efforts inherit a common formulation. Casting unmixing as parameter estimation~\cite{TSP2021ExtendedObjects}, the seminal line of work exploits target sparsity on the imaging plane: each pixel is subdivided into finer sub-pixel cells, an over-complete dictionary is constructed over the resulting grid, and the source distribution is recovered by solving an $\ell_1$-regularized second-order cone program~\cite{Zhang2013Sparse}. This formulation is elegant, but it quietly commits to two consequential design choices. First, it encodes our knowledge of the scene solely as generic mathematical sparsity, agnostic to any scene-specific evidence. Second, it discretizes the continuous space of target locations onto a fixed lattice. Beyond these structural commitments, the resulting optimization is notoriously sensitive to hyperparameter selection~\cite{NIPS2021Hyperparameter}, and the optimal setting drifts with target count and configuration~\cite{TBDATA2017Hyperparameter}, which undermines practical deployment. As we argue below, the two design choices were made for tractability rather than derived from the nature of the problem, and they have since hardened into the defining bottlenecks of the field.

Deep learning offers a natural path forward, having transformed even the most challenging small- and point-target detection tasks~\cite{WACV2021ACM,TGRS2023MultiscaleMultilevel,zhang2025mirsam,wu2026psatiotemporal,deng2025CSPENet,WU2026PR,10824834,10381806,ISPRS2025ISPRS}. In stark contrast, \textbf{\textit{deep learning for CSIST unmixing remains almost entirely unexplored}}. We attribute this stagnation to three barriers, one infrastructural and two paradigmatic:

\textbf{a) Systemic Void in the Research Ecosystem.} Unlike vision domains that thrive on mature infrastructures, CSIST unmixing offers newcomers no public benchmark to train on, no unified metric to compare against, and no baseline implementation to build upon. The consequence goes beyond inconvenience. Data-driven methods advance through benchmark-driven iteration, and in the absence of such infrastructure the field has remained frozen in its optimization-era formulation.

\textbf{b) Inherent Blindness of Under-Constrained Unmixing.} Diffraction renders unmixing an ill-posed, one-to-many inverse problem, in which markedly different source configurations can produce nearly identical blobs. The prevailing remedy, $\ell_1$ sparsity, is semantically blind: it prefers fewer active atoms but knows nothing about how many sources are actually present. Lacking any scene-specific evidence to discriminate among competing explanations, blind reconstruction splits one target into several or fuses several into one, which manifests as persistent false alarms and missed detections. Naively substituting a black-box deep network does not resolve this ambiguity. Without explicit structural priors, it merely relocates the blindness from the regularizer into the network weights.

\textbf{c) Quantization Bottleneck of Grid-Based Separation.} The over-complete dictionary reduces unmixing to deciding which grid cells are occupied, thereby welding predicted coordinates to fixed lattice centers. True target positions, however, live in continuous space and generically fall between grid points, so quantization error is built into the formulation itself. The only escape the paradigm offers is grid densification, an intractable bargain in which computation and memory swell while the error is merely shrunk, never eliminated. Localization precision thus becomes a commodity purchased with computation, at an ever-worsening exchange rate.

Barriers b) and c) share a common root: \textbf{the reduction of CSIST unmixing to a \emph{blind, discrete} sub-pixel separation, a formulation adopted for tractability rather than fidelity to the problem.} A natural question then arises: \textit{can unmixing be upgraded from blind to informed, and from discrete to continuous?} Our answer is affirmative, and it rests on two observations. First, among all latent variables of this inverse problem, the global target count is decisive. The moment the count $N$ is known, the unconstrained one-to-many solution space collapses into a well-structured $N$-source estimation problem. Injecting this single scalar as an explicit semantic prior therefore regularizes the ambiguity at its source, which no generic sparsity penalty can achieve. Second, sub-pixel precision does not require finer grids; it requires abandoning the assumption that targets sit exactly on the grid. If a coarse grid supplies the anchor and a regressor predicts the continuous off-grid offset of each target, then localization accuracy is decoupled from grid resolution. The quantization bottleneck is not mitigated but dissolved.

We actualize this paradigm shift on two levels. At the infrastructure level, we establish the first comprehensive open-source ecosystem for CSIST unmixing, comprising the large-scale CSIST-100K benchmark, a tailored sub-pixel metric suite, and the GrokCSO toolkit, providing the field with the common ground it has lacked. At the algorithmic level, we present DISTA-Net++, a dynamic deep unfolding backbone that inherits the interpretability of sparse reconstruction while learning its transforms and thresholds adaptively. The backbone is armed with two synergistic mechanisms that instantiate our two observations: the Count-Guided Prior (CGP), which predicts the global count and exploits it both as soft modulation over intermediate features and as a hard constraint on final estimates, and the Continuous Coordinate Rectification (CCR), which regresses continuous spatial offsets to free predictions from lattice centers.

Our main contributions are fourfold:
\begin{itemize}
    \item \textbf{A rethinking of the unmixing paradigm.} We identify the blind, discrete sub-pixel separation underlying all prior art as the root cause of the field's stagnation, and advocate its upgrade into an informed, continuous formulation. We expect this perspective to generalize beyond the specific architecture proposed here.
    \item \textbf{The first dedicated research ecosystem.} We build the foundational infrastructure of the field from the ground up: the CSIST-100K benchmark, a sub-pixel evaluation suite including CSO-mAP together with the newly introduced TP-PRMSE and C-ACC metrics, and the open-source GrokCSO toolkit, enabling reproducible and extensible research.
    \item \textbf{DISTA-Net++, an informed and continuous unmixing framework.} Upon a dynamic deep unfolding backbone, CGP converts blind reconstruction into semantic-guided disentanglement by injecting the target count, the most decisive latent variable, as an explicit prior, while CCR achieves genuinely continuous localization by regressing off-grid offsets, severing the tie between precision and grid resolution.
    \item \textbf{Evidence that precision need not be purchased with discretization.} Extensive experiments on CSIST-100K, together with systematic stress tests under varying noise and target densities, show that even under the most economical $3\times$ division, DISTA-Net++ surpasses $7\times$-division state-of-the-art methods by 16.15\% in CSO-mAP and 62.96\% in count accuracy at one-sixth of their computation, directly falsifying the precision-computation exchange assumed by the prevailing paradigm.
\end{itemize}

This work substantially extends our preliminary conference version~\cite{han2025dista} along three axes. \textbf{(i) From blind to informed:} the CGP transforms ambiguity-prone blind reconstruction into a count-guided process. \textbf{(ii) From discrete to continuous:} the CCR breaks the spatial quantization barrier, so that precision gains no longer demand finer sub-pixel divisions. \textbf{(iii) From incremental to transformative:} our new formulation yields an unprecedented performance leap on CSIST-100K, nearly doubling the predecessor's CSO-mAP. Measured by our newly introduced metrics, it achieves a nearly two-fold increase in C-ACC while slashing the TP-PRMSE to roughly one-third. To rigorously validate this breakthrough, we further broaden the experimental protocol with challenging scenarios spanning noise intensities and target densities, yielding a considerably more comprehensive account of robustness and generalization.

\section{Related Work}
\label{sec:related_work}

\subsection{Infrared Small Target Detection}

Fueled by a growing collection of open-source benchmarks~\cite{TGRS2021ALCNet,TGRS2023OSCAR,li2024sm3det}, infrared small target detection has evolved into a mature and advancing field. Since dim targets offer little texture or shape to learn from, the dominant research thread compensates by aggregating evidence across scales and semantic levels~\cite{TGRS2023MultiscaleMultilevel}. Representative efforts include the asymmetric contextual modulation of Dai~\emph{et al.}~\cite{WACV2021ACM}, which exchanges high-level semantics and low-level details through bidirectional attention pathways; the reinforcement-learning-driven pyramid fusion of Wang~\emph{et al.}~\cite{TGRS23RLPGBNet}, which suppresses localized bright clutter with global context boundary attention; and the encoder-decoder design of Tong~\emph{et al.}~\cite{TGRS23MSAFFNet}, which couples atrous spatial pyramid pooling with multiscale edge-aware supervision. Collectively, these methods have pushed the sensitivity of single-frame detection close to its ceiling.

However capable, this work answers only the question of \emph{presence}—whether a suspicious blob exists—remaining silent on the \emph{composition}: the number, sub-pixel locations, and radiation intensities of hidden sources. As argued in Section~\ref{sec:intro}, once targets fall within the diffraction limit, the second question can no longer be answered by any detector operating at the blob level, no matter how sensitive. Our work does not compete with IRSTD methods but picks up precisely where they stop, taking the detected mixture as input and recovering the source-level structure that detection is physically unable to resolve.

% Our work focuses on infrared small targets but differs from previous studies in two main ways. First, while infrared small target detection is a precursor to our study, our primary focus is on the unmixing of closely-spaced infrared small targets. Detecting potentially overlapped targets is essential for subsequent tasks like sub-pixel localization and radiation intensity prediction. Second, our task extends beyond simple detection by precisely locating targets at the sub-pixel level and estimating their radiative intensities, offering a more detailed understanding of target characteristics compared to the previous binary detection approach.

\subsection{Closely-Spaced
Infrared Small Target Unmixing}

Unlike the flourishing detection literature, research on CSIST unmixing is remarkably sparse, and its evolution can be read as a succession of attempts to tame the same underlying ill-posedness. Early studies cast unmixing as parameter estimation~\cite{reagan1993model, lin2012bayesian, macumber2005hierarchical, lin2011qpso} via likelihood or Bayesian inference; while statistically principled, these confront severely non-convex landscapes riddled with local minima and demand prior knowledge rarely available in practice. The field consequently shifted toward sparse reconstruction~\cite{hui2013super, zeng2017infrared}, which discretizes the image plane into sub-pixel cells, builds an over-complete dictionary upon the resulting grid, and recovers source distributions through sparsity-regularized convex programs~\cite{Zhang2013Sparse}. This reformulation trades the non-convexity of parameter estimation for tractability, yet introduces the two structural commitments dissected in Section~\ref{sec:intro}: the scene knowledge is reduced to generic mathematical sparsity that carries no semantic evidence about the actual source configuration, and the solution space is welded to a fixed lattice whose resolution caps localization precision. On top of these commitments, the reconstruction quality remains acutely sensitive to hyperparameter choices~\cite{NIPS2021Hyperparameter,TBDATA2017Hyperparameter}, further eroding deployability.

It is telling that the community has long sensed the insufficiency of blind sparsity. Recent work seeks additional constraints from external hardware, for instance by fusing multi-view observations to disambiguate overlapping configurations~\cite{an2022closely}. The instinct is sound: an ill-posed inverse problem demands more evidence. The remedy, however, looks outward to extra sensors, which inflates system cost and restricts deployment, while overlooking the evidence latent within a single observation. Our work argues that the most decisive piece of such intrinsic evidence is the global target count, and that injecting it as an explicit semantic prior collapses the ambiguous solution space from within, requiring no hardware beyond the original sensor. In parallel, we discard the lattice assumption: rather than approximating continuous positions with ever-finer grids, we regress continuous off-grid offsets directly, so that localization precision is no longer a function of grid resolution. To our knowledge, this constitutes the first deep learning framework for CSIST unmixing, and more importantly, the first to replace the blind, discrete formulation shared by all prior art with an informed, continuous one.

\subsection{Deep Unfolding}

Deep unfolding~\cite{SPM2021AlgorithmUnrolling} occupies a distinctive middle ground between model-driven optimization and end-to-end learning: it inherits the interpretability and convergence structure of iterative algorithms while acquiring the adaptivity of learned parameters. The paradigm originates from LISTA~\cite{ICML2010LISTA}, which reinterprets the iterations of ISTA~\cite{Daubechies2004ISTA} as layers of a feed-forward network and attains comparable accuracy with far fewer iterations. Subsequent milestones broadened the recipe to other solvers and domains: ADMM-Net~\cite{NIPS2016ADMMNet} unfolds the alternating direction method of multipliers for compressive MRI reconstruction, and ISTA-Net~\cite{CVPR2018ISTANet} learns the proximal mapping end-to-end for natural image compressive sensing. The paradigm has since permeated diverse inverse problems in imaging, including blind deblurring via unfolded gradient-domain total variation~\cite{TCI2020EfficientDeblurring}, super-resolution with trainable transforms embedded in the DCT framework~\cite{guo2019adaptive}, and ultrasound clutter suppression through unfolded robust PCA~\cite{TMI2020DeepUnfoldedRPCA}.

Given that the dominant formulation of CSIST unmixing is a sparse reconstruction problem, deep unfolding is its natural learning-based successor, yet this connection has never been exploited. We bridge this gap with a design choice motivated by the physics of the task. Existing unfolding networks~\cite{CVPR2018ISTANet,ICME2021ISTANETPlusPlus} freeze their transforms and thresholds after training, implicitly assuming a homogeneous input distribution. CSIST scenes violate this assumption by nature: unmixing difficulty varies drastically with target count, spacing, and noise level, so a single static configuration is necessarily a compromise. DISTA-Net++ therefore generates its proximal mapping weights dynamically, conditioned on the input observation, allowing each mixture to be reconstructed under parameters tailored to its own difficulty. The unfolding backbone thus serves not as an off-the-shelf borrowing but as the structural carrier upon which our count-guided prior and continuous rectification operate.

% !TEX root = ../main.tex
% \bibliography{../reference.bib}

\begin{figure*}[!t]
  \centering
\includegraphics[width=.92\textwidth]{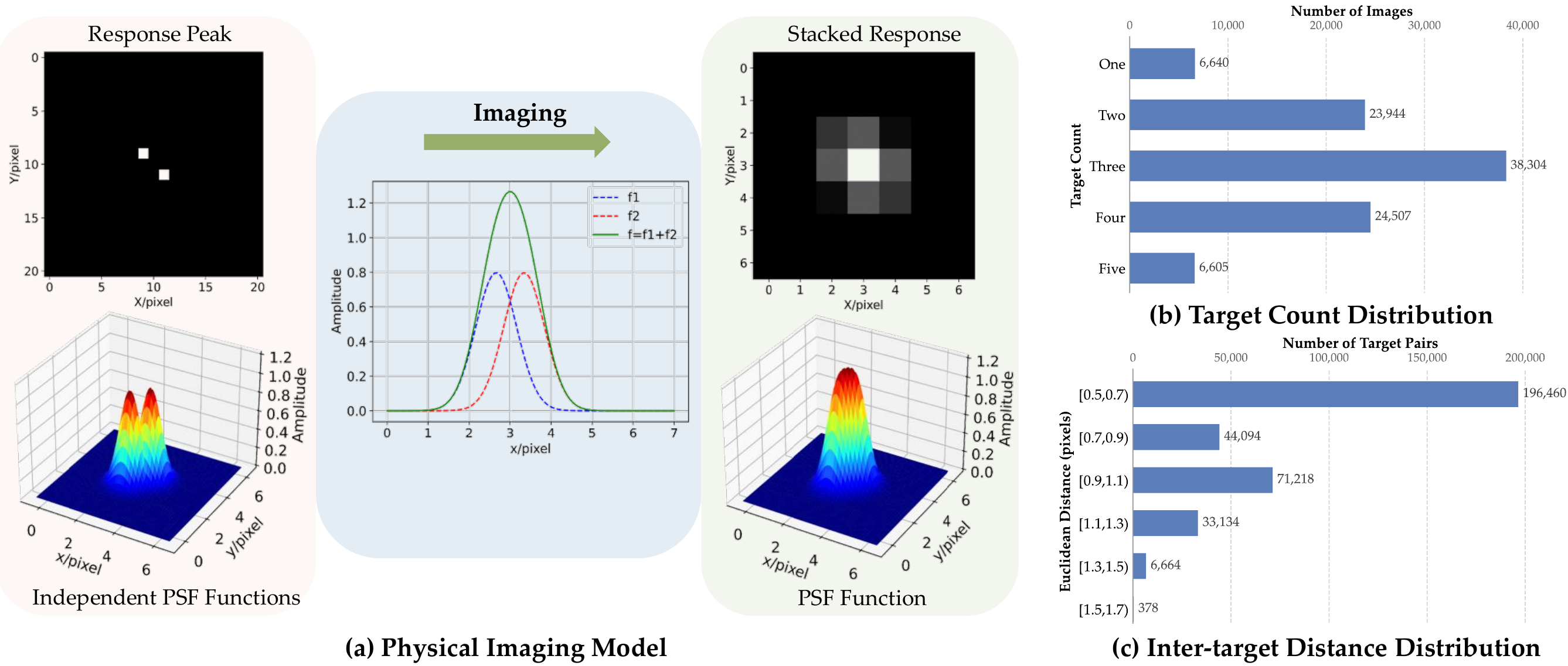}

\caption{\textbf{Overview of the CSIST-100K dataset and the multi-target imaging process.}
(a) \textbf{Physical Forward Imaging Model.} Illustration of the energy overlap process: (Left) independent sub-pixel point sources and their PSFs; (Middle) a 1D cross-section demonstrating cumulative linear superposition ($f = f_1 + f_2$); (Right) the final stacked focal-plane response where closely spaced targets merge into a visually indivisible peak.
(b) \textbf{Target Count Distribution.} The statistical distribution of images across varying target counts (1 to 5 targets).
(c) \textbf{Inter-target Distance Distribution.} The distribution of Euclidean distances (in pixels) between target pairs. The vast majority fall within the extremely close range of $[0.5, 0.7)$ pixels, inducing severe spatial overlap well below the classical resolution limit.}
  \label{fig:dataset_overview}
\end{figure*}

% \section{CSIST Benchmark, Metric, and Toolkit} 
\section{A Research Ecosystem for CSIST Unmixing} 
\label{sec:ecosystem}

% \begin{figure*}[!t]
%     \centering
%     \includegraphics[width=.92\textwidth]{figure/paper/eco-1.pdf}
%     \caption{
%     Sample visualizations from the CSIST-100K dataset: The top row shows 1 to 5 overlapping targets, and the following rows display reconstruction map for sub-pixel division factors of $3 \times$, $5 \times$, and $7 \times$.
%     }
%     \label{fig:gallery}
% \end{figure*}

A task can only advance as far as its infrastructure allows. For CSIST unmixing to grow from scattered attempts into a reproducible research field, three foundations must be laid: a benchmark that supplies exact source-level ground truth, an evaluation protocol that measures what the task actually demands, and an open toolkit that turns isolated results into cumulative progress. None of the three exists in the current literature. This section constructs each in turn.

\subsection{The CSIST-100K Benchmark}
\label{subsec:dataset}

Building a benchmark for CSIST unmixing confronts an obstacle that is epistemic rather than logistic. The labels this task requires, namely sub-pixel positions and radiant intensities of individual sources, describe precisely the information that the sensor cannot resolve. No human annotator, however expert, can recover source-level truth from an observed blob whose ambiguity is imposed by physics itself. Real-data annotation is therefore not merely expensive but impossible in principle. Fortunately, the same physics that forbids annotation also enables synthesis: the forward imaging process of point sources is governed by diffraction optics that is understood to first principles, so a simulated benchmark grounded in this process provides ground truth that is not approximately correct but exactly correct. Simulation here is not a compromise forced by data scarcity; it is the only principled route to supervision for this task.

We accordingly model the imaging process as follows. When a point source is imaged through a circular aperture, the focal plane exhibits an Airy pattern, with a central lobe carrying 84\% of the total energy surrounded by concentric diffraction rings. This response is conventionally approximated by a Gaussian point spread function (PSF), whose standard deviation is determined by the sensor's focal ratio and detection band. For remote objects, each target acts as an ideal point source whose Airy radius equals $1.22 \lambda / D$, where $\lambda$ denotes the wavelength and $D$ the aperture diameter. This radius, equivalent to $1.9\sigma$ of the Gaussian PSF, defines one Rayleigh unit, the classical resolution limit of the sensor. Under multi-target imaging, the intensity recorded at each pixel is the linear superposition of all source responses, as illustrated in Fig.~\ref{fig:dataset_overview}(a).

Upon this physical model we construct CSIST-100K. Each sample places 1 to 5 targets on an $11 \times 11$ pixel grid with $\sigma_{\mathrm{PSF}} = 0.5$ pixels, where every target carries a continuous sub-pixel position and a radiant intensity drawn from 220 to 250 units. Pairwise separations are permitted to fall far below one Rayleigh unit, down to 0.52 Rayleigh units, so that a substantial fraction of samples lies squarely in the regime where classical resolution fails and unmixing becomes the only recourse. The dataset comprises 100,000 samples, split into 80,000 for training and 10,000 each for validation and testing. The statistical distributions of target counts and pairwise separations are reported in Fig.~\ref{fig:dataset_overview}(b) and (c). Beyond this standard setting, we further extend the benchmark to stress conditions with severe background noise and saturated target density, deferring details to Sec.~\ref{subsec:sota}, so that generalization claims can be tested rather than assumed.

\subsection{Task-Aligned Evaluation Metrics}
\label{subsec:metric}

Evaluation deserves the same first-principles scrutiny as data. The prevailing metrics of object detection are built upon bounding-box overlap, and overlap-based matching presupposes that neighboring targets occupy distinguishable spatial extents. This presupposition is exactly what diffraction destroys: below the Rayleigh limit, the boxes of adjacent targets collapse into near-identical regions, and IoU ceases to carry any discriminative signal. A task defined in the continuous source domain must be evaluated in that same domain. We therefore design a metric suite whose structure mirrors the structure of the task itself: a primary metric that scores the joint recovery of count, position, and intensity, complemented by orthogonal probes that isolate each capability for diagnosis.

\subsubsection{CSO-mAP}
As the primary metric, Closely-Spaced Objects mean Average Precision 
(CSO-mAP) replaces box overlap with distance-based matching in continuous 
coordinates. A prediction $\hat{t}_j$ is matched to a ground-truth target 
$t_i$ through the indicator
\begin{equation}
\mathbbm{1}_k\left(\hat{t}_j, t_i\right)=\left\{\begin{array}{lc}
1, & \text { if } d\left(\hat{t}_j, t_i\right) < \delta_k, \\
0, & \text { otherwise, }
\end{array}\right.
\label{eq:cso-mAP}
\end{equation}
where $\delta_k \in \{0.05, 0.1, 0.15, 0.2, 0.25\}$ pixels 
($k = 1, \dots, 5$) is a series of thresholds of increasing leniency. 
Following the standard average precision (AP) paradigm, predictions are 
ranked by their intensity confidence to trace the precision-recall curve, 
the area under which yields the AP at each $\delta_k$. CSO-mAP averages 
these AP values across all thresholds, thereby summarizing unmixing 
quality over a spectrum of localization stringencies within a single 
number.

\subsubsection{TP-PRMSE}
CSO-mAP entangles two abilities: finding the correct targets and locating 
them precisely. To isolate the latter, True Positive Position Root Mean 
Square Error (TP-PRMSE) computes the positional error over successfully 
matched pairs only,
\begin{equation}
\text{TP-PRMSE} = \sqrt{\frac{1}{|\mathcal{M}|} \sum_{(\hat{t}_j, t_i) \in \mathcal{M}} d\left(\hat{t}_j, t_i\right)^2},
\label{eq:tprmse}
\end{equation}
where $\mathcal{M}$ is the set of matched prediction-target pairs 
established during CSO-mAP evaluation. By conditioning on correct matches, 
this metric answers a question that CSO-mAP alone cannot: given that a 
target is found, how precisely is it pinned down?

\subsubsection{C-ACC}
Resolving the overlap ambiguity manifests most directly in predicting how 
many sources a blob contains. Count Accuracy (C-ACC) probes this 
capability at the image level through exact count matching,
\begin{equation}
\text{C-ACC} = \frac{1}{N} \sum_{n=1}^{N} \mathbbm{1}\left(\hat{C}_n = C_n\right),
\label{eq:cacc}
\end{equation}
where $N$ is the number of evaluated samples, and $\hat{C}_n$ and $C_n$ 
denote the predicted and ground-truth counts of the $n$-th sample. The 
indicator rewards only exact agreement, so both over-counting and 
under-counting are penalized without partial credit. Given the central 
role that the target count plays in our framework, this metric also 
serves as a direct diagnostic of the count-guided prior itself.

\subsubsection{PSNR and CSO-SSIM}
Finally, we evaluate the radiometric fidelity of the reconstructed spatial 
response maps using Peak Signal-to-Noise Ratio (PSNR) and a tailored variant 
of the Structural Similarity Index Measure (SSIM). Standard global SSIM is 
poorly suited for the inherently sparse CSIST spatial response maps, as the 
vast empty background dominates the score and masks structural errors within 
actual sources. We therefore introduce Closely-Spaced Objects Structural 
Similarity (CSO-SSIM), which stringently penalizes localized distortions by 
computing similarity exclusively within $3 \times 3$ patches centered on ground-truth 
targets, effectively discarding the uninformative background. High scores 
under these measures require not only correct counts and positions but also 
accurate radiant intensities, thereby completing the coverage of all three 
quantities the task demands.

\subsection{The GrokCSO Toolkit}
\label{subsec:toolkit}

A nascent field is most vulnerable not to hard problems but to fragmentation. When every group re-implements baselines from scratch, results cease to be comparable, and effort that should compound instead dissipates. Mature vision tasks avoid this fate through shared platforms such as MMDetection and GluonCV, yet these frameworks offer little shelter here: their core abstractions of anchors, boxes, and non-maximum suppression are native to extended objects and simply do not speak the language of continuous point sources. CSIST unmixing needs a platform whose primitives match its own.

GrokCSO fills this role. Built on PyTorch, it provides unified implementations of representative unmixing algorithms together with pre-trained models, training scripts, and full logs, so that every result reported in this paper can be reproduced from a single command. Its modular design decouples backbones, unfolding structures, and prediction heads, allowing new architectures to be composed and benchmarked with minimal effort. Task-specific infrastructure, including dataset loaders for CSIST-100K, augmentation pipelines respecting sub-pixel geometry, and reference implementations of all metrics in Sec.~\ref{subsec:metric}, is provided out of the box. We release GrokCSO not as a supplement to this paper but as an invitation: by lowering the entry barrier for the community, we hope the benchmark, the metrics, and the toolkit together turn CSIST unmixing from a problem we study into a field others can build upon.

\section{DISTA-Net++: Informed and Continuous Unmixing}
\label{sec:DISTA_PLUS}

\begin{figure}[htbp]
\centering
\includegraphics[width=.42\textwidth]{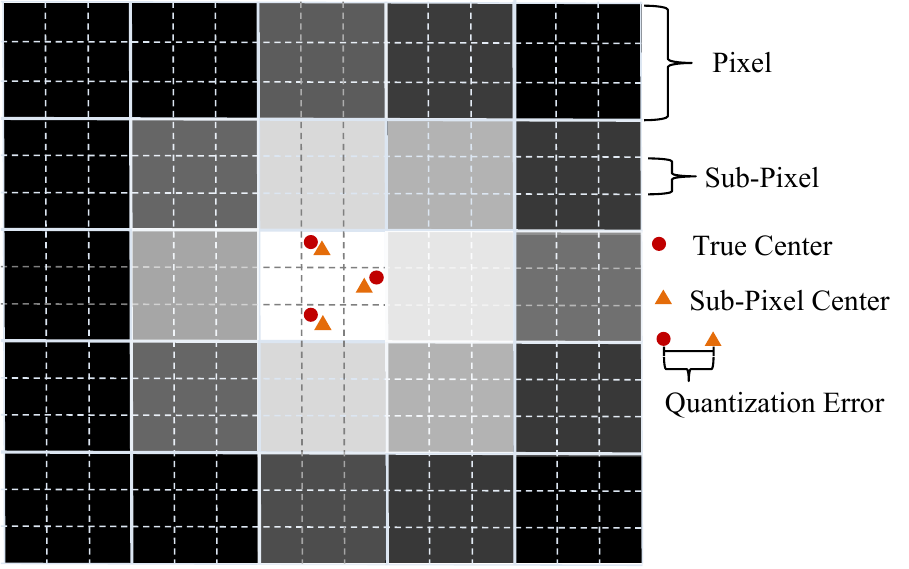}
\caption{Division of each pixel into an $n\times n$ grid of sub-pixels, representing potential target positions.}
\label{fig:subpixels}
\end{figure}

\begin{figure*}[!t]
    \centering
    \includegraphics[width=.92\textwidth]{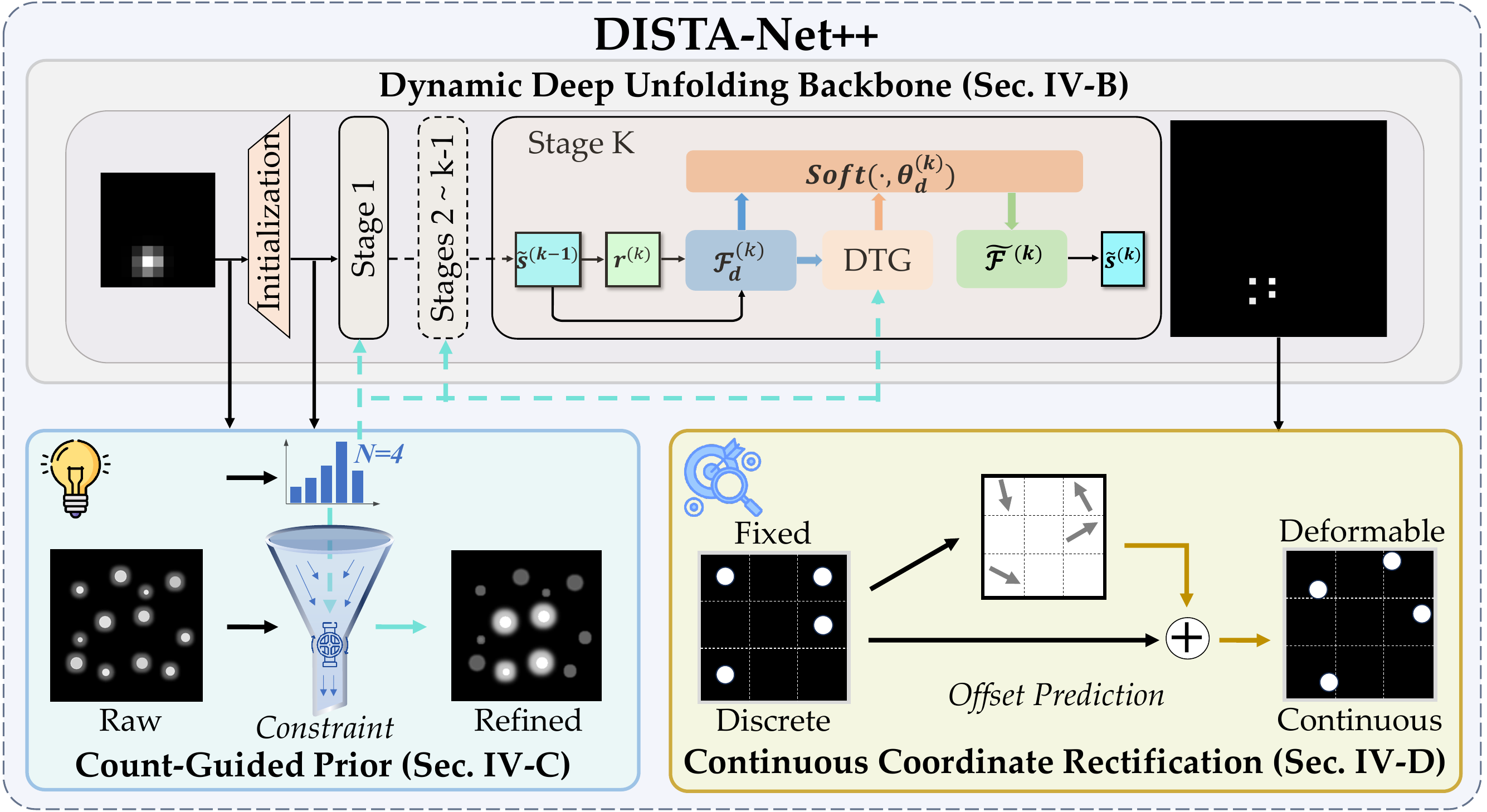}
    \caption{\textbf{Overall architecture of the proposed DISTA-Net++.} The framework is intrinsically driven by three synergistic components: (a) \textbf{Dynamic Deep Unfolding Backbone} (top) that adaptively extracts features via a Dynamic Transform (DT) module and computes stage-wise soft-thresholds via a Dynamic Thresholding Generator (DTG) to handle sensitive signal variations; (b) \textbf{Count-Guided Prior} module (bottom left) that leverages global target count to semantically modulate features and suppress overlapping ambiguity; and (c) \textbf{Continuous Coordinate Rectification} mechanism (bottom right) that regresses sub-pixel offsets to refine discrete grid estimations into precise continuous spatial localization.}
    \label{fig:DISTA-Net++}
\end{figure*}

DISTA-Net++ is not a collection of modules but the architectural embodiment of the two diagnoses established in Sec.~\ref{sec:intro}. The first diagnosis holds that existing unmixing is blind: lacking any global semantic anchor, the network must disambiguate overlapping sources from local intensity patterns alone, a task that diffraction renders fundamentally underdetermined. The second holds that existing unmixing is discrete: by casting localization as on-grid classification, it trades localization fidelity for separation capability and inherits an irreducible quantization error.

The design of DISTA-Net++ follows a simple logic. We retain deep unfolding as the physics-grounded backbone, since it alone preserves the interpretable correspondence between network stages and optimization iterations, and repair its two congenital defects: a Count-Guided Prior converts blind inference into informed inference, and a Continuous Coordinate Rectification releases localization from the grid into the continuous domain. The two remedies converge at inference through a count-constrained extraction procedure, and are jointly realized at training time through a coupled objective. The overall architecture is depicted in Fig.~\ref{fig:DISTA-Net++}. This section develops each component in order.

\subsection{Problem Formulation}
\label{sec:framework}

\noindent \textbf{CSIST Imaging Model.}
Given the vast distance between targets and the infrared detector, each target is well approximated as a point source. Optical diffraction spreads its energy across adjacent pixels, a process conventionally modeled by a two-dimensional Gaussian point spread function (PSF)~\cite{AA2021PSF}:
\begin{equation}
p(x, y)=\frac{1}{2 \pi \sigma_{\mathrm{PSF}}^2} \exp \left[-\frac{\left(x-x_t\right)^2+\left(y-y_t\right)^2}{2 \sigma_{\mathrm{PSF}}^2}\right],
\label{eq:psf}
\end{equation}
where $\sigma_{\mathrm{PSF}}^2$ is the diffusion variance and $(x_t, y_t)$ denotes the target position. On a focal plane of $U \times V$ pixels, each pixel integrates the PSF within its boundaries:
\begin{equation}
g_{i, j}\left(x_t, y_t\right)=\int_{x_{i, j} - D/2}^{x_{i, j}+ D/2} \int_{y_{i, j}- D/2}^{y_{i, j}+ D/2} p(x, y) \,\mathrm{d} x \,\mathrm{d} y,
\label{eq:response}
\end{equation}
where $(x_{i,j}, y_{i,j})$ is the pixel center and $D$ the pixel width. Stacking all pixels yields the vectorized measurement model
\begin{equation}
\mathbf{z}=\mathbf{G}(x, y)\, \mathbf{s}+\mathbf{n},
\label{eq:measurement}
\end{equation}
where $\mathbf{G}(x, y)$ is the steering matrix, $\mathbf{s}$ collects the target intensities, and $\mathbf{n}$ is Gaussian white noise. The unmixing task is thus a blind inverse problem: neither the number of columns that are active in $\mathbf{G}$, nor their continuous positions, nor their amplitudes are known a priori.

\noindent \textbf{Sparse Reconstruction on a Sub-Pixel Grid.}
\label{subsec:proximal}
Since the continuous positions render Eq.~\eqref{eq:measurement} nonlinear in $(x_t, y_t)$, the classical remedy is to linearize by discretization. As shown in Fig.~\ref{fig:subpixels}, each pixel is partitioned into an $n \times n$ grid, producing a candidate position set $\Omega=\left\{(x_l, y_l)\right\}_{l=1}^{L}$ with $L=UVn^2$ sub-pixel cells, where the true targets occupy a sparse subset and the worst-case quantization deviation is $\sqrt{2}D/2n$. The measurement model then becomes linear over $\Omega$ as $\mathbf{z}=\mathbf{G}(\Omega)\, \tilde{\mathbf{s}}+\mathbf{n}$, where $\mathbf{G}(\Omega)$ stacks the steering vectors of all candidate cells and $\tilde{\mathbf{s}} \in \mathbb{R}^{L}$ is sparse with $L \gg UV$. Unmixing is accordingly cast as $\ell_1$-regularized sparse recovery:
\begin{equation}
\min_{\tilde{\mathbf{s}}} \Vert \mathbf{z}-\mathbf{G}(\Omega)\, \tilde{\mathbf{s}}\Vert_2^2+\lambda\Vert\tilde{\mathbf{s}}\Vert_1,
\label{eq:reconstruction}
\end{equation}
whose solution simultaneously encodes all three target attributes: the support of $\tilde{\mathbf{s}}$ gives the count, the corresponding cells in $\Omega$ give the positions, and the nonzero amplitudes give the intensities. We emphasize a structural consequence of this formulation that motivates Sec.~\ref{subsec:coor}: once the grid is fixed, localization accuracy is bounded by quantization, and this bound cannot be lifted by any solver within the formulation.

\noindent \textbf{From ISTA to Unfolding.}
The classical solver for Eq.~\eqref{eq:reconstruction} is ISTA~\cite{Daubechies2004ISTA}, which alternates a gradient descent step on the data fidelity with a proximal step on the sparsity prior:
\begin{align}
  \mathbf{r}^{(k)} &= \tilde{\mathbf{s}}^{(k-1)} - \rho\, \mathbf{G}^{\top}\big(\mathbf{G} \tilde{\mathbf{s}}^{(k-1)} - \mathbf{z}\big), \label{eq:update-r} \\
  \tilde{\mathbf{s}}^{(k)} &= \underset{\tilde{\mathbf{s}}}{\arg \min }\, \frac{1}{2}\big\|\tilde{\mathbf{s}} - \mathbf{r}^{(k)}\big\|_2^2 + \lambda\|\boldsymbol{\Psi} \tilde{\mathbf{s}}\|_1, \label{eq:update-x_1}
\end{align}
where $\boldsymbol{\Psi}$ is a sparsifying transform, $k$ indexes iterations, and $\rho$ is the step size. The second step is the proximal mapping $\operatorname{prox}_{\lambda \phi}(\mathbf{r})=\arg \min_{\tilde{\mathbf{s}}} \frac{1}{2}\|\tilde{\mathbf{s}}-\mathbf{r}\|_2^2+\lambda \phi(\tilde{\mathbf{s}})$. ISTA admits closed-form proximal solutions only for orthogonal transforms and converges slowly. Deep unfolding, exemplified by ISTA-Net~\cite{CVPR2018ISTANet}, truncates the iteration to a fixed number of stages and replaces the handcrafted $\boldsymbol{\Psi}$ with a learnable nonlinear transform $\mathcal{F}(\cdot)$:
\begin{equation}
  \tilde{\mathbf{s}}^{(k)} = \underset{\tilde{\mathbf{s}}}{\arg \min }\, \frac{1}{2}\big\| \mathcal{F}\big( \tilde{\mathbf{s}} \big) - \mathcal{F}\big( \mathbf{r}^{(k)} \big)\big\|_2^2 + \theta\|\mathcal{F}\big( \tilde{\mathbf{s}} \big)\|_1, \label{eq:update-x_2}
\end{equation}
with $\theta$ learned from data. This marriage of physics and learning is the right foundation for CSIST unmixing, yet it carries a subtle mismatch with our task. Once trained, both the transform and the threshold are frozen, whereas CSIST observations vary drastically from sample to sample in target density and overlap severity. A transform tuned for isolated targets underserves dense clusters, and a single threshold is inevitably too aggressive for weak overlapping signals or too permissive for clean backgrounds. The backbone we develop next removes this rigidity.

\subsection{Dynamic Deep Unfolding Backbone}
\label{sec:architecture}

\begin{figure*}[htbp]
    \centering
    \includegraphics[width=.92\textwidth]{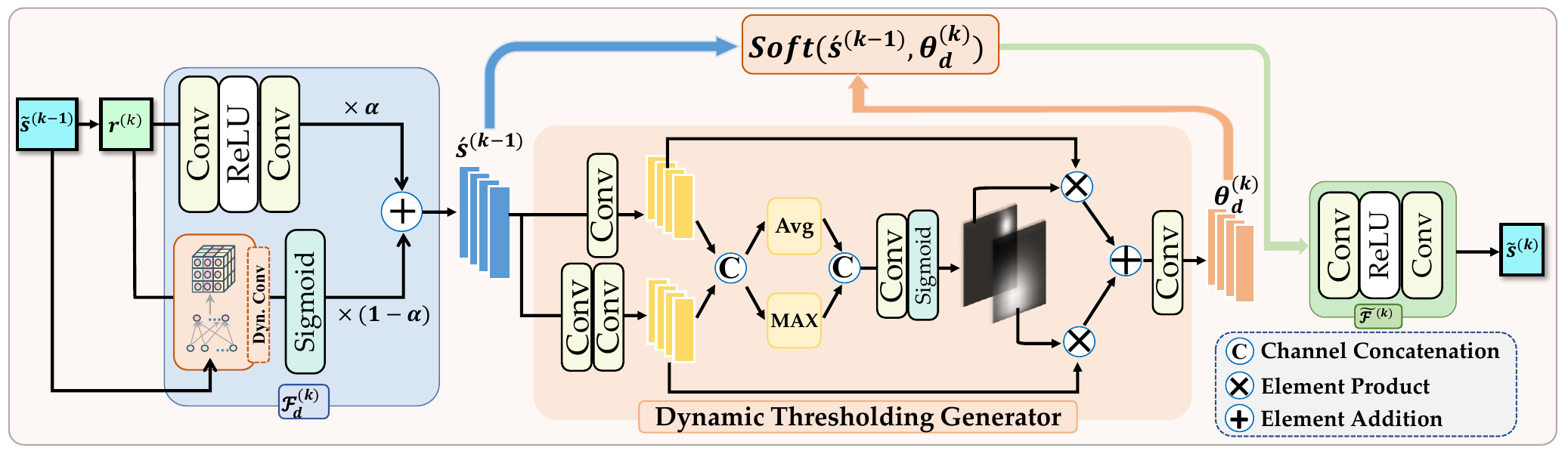}
    \caption{\textbf{Architecture of the $k$-th stage in the dynamic deep unfolding backbone.} Each stage contains three main components: a dual-branch dynamic transform module ($\mathcal{F}_d^{(k)}$) for feature extraction, a Dynamic Thresholding Generator (DTG) for feature refinement, and an inverse transform module ($\tilde{\mathcal{F}}^{(k)}$) for reconstruction. In the full DISTA-Net++ framework, the DTG is further modulated by a global semantic count prior (omitted in this diagram for visual clarity) to adaptively regulate the thresholds.}
    \label{fig:DDUN}
\end{figure*}

The guiding principle of our backbone, detailed in Fig.~\ref{fig:DDUN}, is to promote the two frozen ingredients of unfolding, the transform and the threshold, from fixed parameters to functions of the input. Each stage $k$ retains the unfolded structure of Eq.~\eqref{eq:update-x_2}, computing $\tilde{\mathbf{s}}^{(k)} = \tilde{\mathcal{F}}^{(k)}\big(\operatorname{Soft}\big(\mathcal{F}_d^{(k)}(\mathbf{r}^{(k)}),\, \theta_d^{(k)}\big)\big)$, where $\mathbf{r}^{(k)}$ follows Eq.~\eqref{eq:update-r}, $\operatorname{Soft}(\cdot, \theta_d)$ is soft-thresholding, and $\tilde{\mathcal{F}}^{(k)}$ is a left inverse satisfying $\tilde{\mathcal{F}}^{(k)} \circ \mathcal{F}_d^{(k)} = \mathbf{I}$ without requiring structural symmetry. Following ISTA-Net~\cite{CVPR2018ISTANet}, the recurrence begins with an initial estimate $\tilde{\mathbf{s}}^{(0)}$ computed via an optimal linear projection of the input $\mathbf{z}$. What distinguishes our backbone is that both $\mathcal{F}_d^{(k)}$ and $\theta_d^{(k)}$ are generated conditionally on the observation itself.

\noindent \textbf{Dynamic Transform.}
The transform $\mathcal{F}_d^{(k)}$ adopts a dual-branch design. A static branch applies a standard Conv-ReLU-Conv mapping to $\mathbf{r}^{(k)}$, providing a stable feature basis shared across all inputs. An auxiliary dynamic branch then adapts this basis to the sample at hand: a global context vector is extracted from the previous estimate $\tilde{\mathbf{s}}^{(k-1)}$ via global average pooling and passed through a lightweight two-layer network of $1 \times 1$ convolutions to generate sample-specific convolutional kernels $W = \sigma\big(\mathcal{C}_2\big(\delta\big(\mathcal{C}_1\big(P_{\text{avg}}(\tilde{\mathbf{s}}^{(k-1)})\big)\big)\big)\big)$, where $\sigma(\cdot)$ and $\delta(\cdot)$ denote the sigmoid and ReLU functions, and $W$ is a batch of $m \times m$ kernels, one per sample. These kernels are applied to $\mathbf{r}^{(k)}$ through an efficient grouped convolution $w_r = \mathcal{C}_{\mathrm{dynamic}}\big(\mathbf{r}^{(k)},\, W\big)$, and the two branches are fused by a learnable gate $\alpha \in [0,1]$ as $\acute{\mathbf{s}}^{(k)} = \alpha \cdot \mathcal{C}_2\big(\delta\big(\mathcal{C}_1(\mathbf{r}^{(k)})\big)\big) + (1-\alpha)\cdot \sigma(w_r)$. The rationale is that the previous estimate $\tilde{\mathbf{s}}^{(k-1)}$ already carries a coarse hypothesis about where and how densely targets reside; conditioning the kernels on it allows each stage to reshape its feature extraction according to the evolving reconstruction, rather than applying one pattern to all scenes.

\noindent \textbf{Dynamic Soft-Thresholding.}
The threshold plays an even more delicate role, as it directly arbitrates which activations survive as targets and which are suppressed as noise. We replace the fixed $\theta$ with a \textit{Dynamic Thresholding Generator} (DTG) that produces a spatially varying threshold map from the current features. As shown in Fig.~\ref{fig:DDUN}, the input feature $\mathcal{F}_d^{(k)}$ is processed by two parallel branches, a $1 \times 1$ convolution and a $3 \times 3$ grouped convolution followed by a $1 \times 1$ convolution, yielding multi-scale maps $\tilde{U}_1$ and $\tilde{U}_2$ that are concatenated into $\tilde{U} = [\tilde{U}_1; \tilde{U}_2]$. Spatial saliency is then distilled through parallel average- and max-pooling, and a convolution $\mathcal{C}^{2 \to N}$ with sigmoid activation produces selective masks $\tilde{SA} = \sigma \big( \mathcal{C}^{2 \to N} \big( [P_{\text{avg}}(\tilde{U}); P_{\text{max}}(\tilde{U})] \big) \big)$. The dynamic threshold is obtained by modulating the input feature with the mask-weighted multi-scale aggregation:
\begin{equation}
    \theta_d = \mathcal{F}_d^{(k)} \cdot \mathcal{C} \left( \sum_{i=1}^{N} (\tilde{SA})_i \cdot \tilde{U}_i \right).
    \label{eq:theta}
\end{equation}
In effect, the shrinkage strength is no longer a global constant but a per-location decision informed by local spatial context, which is precisely what densely overlapped clusters demand.

\subsection{Count-Guided Prior: From Blind to Informed}
\label{sec:count_prior}

The backbone above, however adaptive, still reasons blindly. Its every decision rests on local intensity patterns, yet diffraction makes such patterns intrinsically ambiguous: a bright blob may be one strong source or several weak ones, and no pixel-level evidence can settle the question. Our key observation is that the ambiguity is largely resolved once a single global quantity is known, namely how many sources the scene contains. The count is the most compact semantic summary of a CSIST scene, and knowing it collapses the feasible solution set of the inverse problem dramatically. We therefore extract this quantity explicitly and inject it throughout the unmixing process, turning blind inference into informed inference.

\noindent \textbf{Target Counting Network.}
We instantiate the prior with a lightweight counting head $\mathcal{F}_\text{count}$ that predicts the target count as a multi-class classification. Given the observation $\mathbf{Z}$, hierarchical features are extracted by two convolutional blocks $\mathbf{F}_2 = \downarrow\big(\delta\big(\mathcal{C}\big(\downarrow\big(\delta\big(\mathcal{C}(\mathbf{Z})\big)\big)\big)\big)\big)$, where $\downarrow$ denotes max pooling. After flattening, two fully connected layers produce the count embedding $\mathbf{c} \in \mathbb{R}^{64}$ and the classification logits $\mathbf{N}$ as $\mathbf{c} = \phi\big(\delta\big(\mathcal{L}(\mathbf{F}_2)\big)\big)$ and $\mathbf{N} = \mathcal{L}(\mathbf{c})$, with $\mathcal{L}$ a linear layer and $\phi$ dropout. Notably, the prior acts through two pathways: the continuous embedding $\mathbf{c}$ conditions the reconstruction below, while the discrete prediction $\hat{N}$ constrains the final extraction in Sec.~\ref{sec:inference}.

\begin{figure}[t!]
    \centering
    \includegraphics[width=0.98\linewidth]{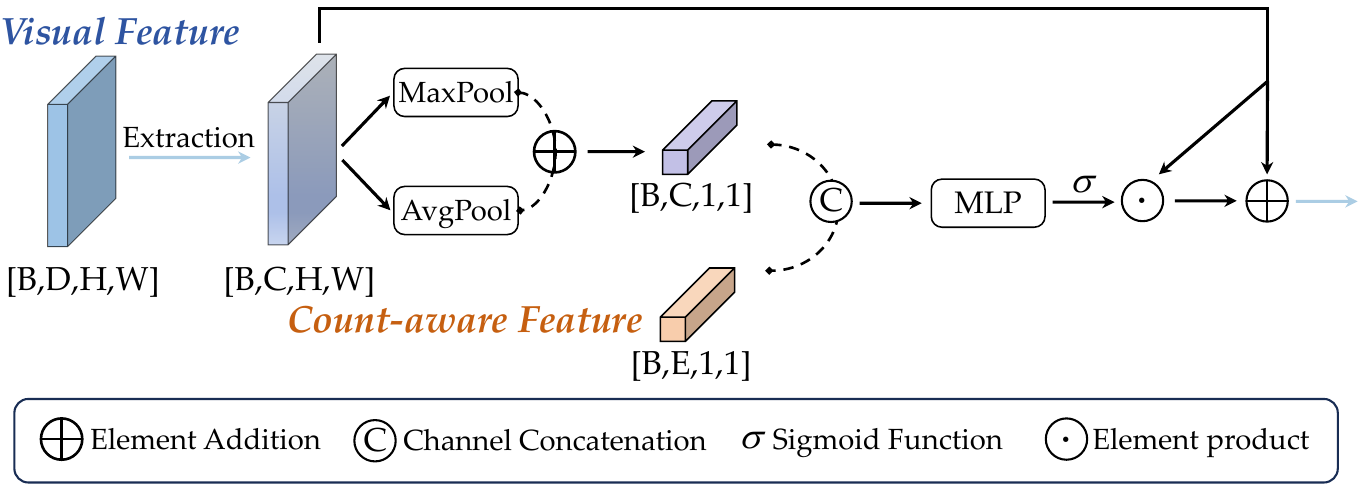}
    \caption{\textbf{Detailed architecture of the Count-Aware Modulation (CAM) operation.} It effectively integrates the spatial visual features with the semantic count feature to produce channel-wise modulation weights.}
    \label{fig:cam}
\end{figure}

\noindent \textbf{Count-Guided Modulating.}
The most natural entry point for the count prior is the threshold, since the appropriate shrinkage strength depends directly on how many sources are present: a scene with few targets calls for aggressive suppression of spurious activations, whereas a dense cluster demands leniency to preserve weak overlapping signals. We formalize this coupling as a \textit{Count-Aware Modulation} (CAM) operator $\Phi_{\text{CAM}}(\mathbf{V}, \mathbf{c})$. Given a visual feature $\mathbf{V}$, CAM first extracts an intermediate spatial representation $\mathbf{X} = \mathcal{C}_2(\delta(\mathcal{C}_1(\mathbf{V})))$, aggregates its global context via average- and max-pooling, concatenates the result with the count embedding $\mathbf{c}$, and produces channel-wise modulation weights through a two-layer MLP $\psi(\cdot)$ with SiLU activation, applied as a sigmoid-gated residual:
\begin{equation}
    \Phi_{\text{CAM}}(\mathbf{V}, \mathbf{c}) = \mathbf{X} + \mathbf{X} \odot \sigma\big(\psi\big([P_{\text{avg}}(\mathbf{X}) + P_{\text{max}}(\mathbf{X}) ;\, \mathbf{c}]\big)\big),
    \label{eq:cam}
\end{equation}
where $[\cdot\,;\cdot]$ denotes channel-wise concatenation and $\odot$ element-wise multiplication. The count-guided dynamic threshold is then obtained by substituting the count-modulated representation into Eq.~\eqref{eq:theta}:
\begin{equation}
    \hat{\theta}_d = \Phi_{\text{CAM}}\big(\mathcal{F}_d^{(k)}, \mathbf{c}\big) \cdot \mathcal{C} \left( \sum_{i=1}^{N} (\tilde{SA})_i \cdot \tilde{U}_i \right).
    \label{eq:theta_count}
\end{equation}
Through this mechanism the network raises thresholds to suppress noise when the scene is sparse, and lowers them to protect weak signals when the scene is dense, with the decision anchored to an explicit global semantic rather than guessed from local evidence.

\subsection{Continuous Coordinate Rectification: From Discrete to Continuous}
\label{subsec:coor}

The second defect requires a different cure. Recall from Sec.~\ref{sec:framework} that grid-based formulations bound localization accuracy by the quantization step $\sqrt{2}D/2n$, and that refining the grid is a losing bargain: the dictionary dimension grows quadratically in $n$, adjacent steering vectors become nearly collinear, and the recovery problem grows harder precisely as the grid grows finer. Our resolution is to stop asking the grid to do what it cannot. We let the grid do what it does well, separating sources and assigning them to cells, and delegate the residual sub-cell displacement to a dedicated regression in the continuous domain. Localization accuracy is thereby decoupled from grid resolution altogether.

Concretely, we append a lightweight offset regression head $\mathcal{R}$ to the terminus of the backbone. To supply sufficient spatial reference, a shallow feature $\mathbf{F}_0$ is extracted from the initial estimate $\tilde{\mathbf{s}}^{(0)}$ via a two-layer convolution block, and $\mathcal{R}$ decodes the concatenation of the final reconstruction $\tilde{\mathbf{s}}^{(N)}$ and $\mathbf{F}_0$ into a continuous offset field $\mathbf{\Delta} = \mathcal{R}\big([\tilde{\mathbf{s}}^{(N)} ;\, \mathbf{F}_0]\big) \in [-1, 1]^{2 \times H \times W}$, where each spatial element $\mathbf{\Delta}_{x',y'} = (\delta x', \delta y')$ predicts the sub-cell displacement of a potential target anchored at grid index $(x', y')$. The head $\mathcal{R}$ is instantiated as an expansion-compression block of $3 \times 3$ Conv-ReLU-Conv layers whose hidden dimension is expanded sixteenfold, granting sufficient capacity for spatial transformation learning before projection into the two-dimensional offset space. A terminal $\tanh$ activation serves two purposes: it bounds the offsets within $[-1, 1]$, acting as a structural regularizer against excessive shifts and training instability, and its symmetric range permits bidirectional adjustment along both axes, which arbitrary-direction sub-pixel localization requires. The final continuous coordinates are recovered as
\begin{equation}
    x = x' + \delta x', \quad y = y' + \delta y'.
    \label{eq:pos_refine}
\end{equation}
With this single addition, the estimator escapes the lattice that has confined grid-based unmixing since its inception, at a computational cost that is negligible relative to the backbone.

\subsection{Count-Constrained Inference}
\label{sec:inference}

The two priors developed above converge at inference. Conventional extraction pipelines select local maxima above a fixed brightness threshold, a procedure that is brittle in exactly the regime we target: within severely blurred clusters, a threshold low enough to retain weak sources also admits diffraction sidelobes as false alarms, while a threshold high enough to reject sidelobes discards genuine targets. The predicted count $\hat{N}$ offers a principled escape, since it specifies in advance how many detections the scene should yield.

Our inference proceeds as follows. Candidates are first gathered from the reconstructed response map by a permissive baseline threshold, then rectified to continuous coordinates $\mathbf{p}_i \in \mathbb{R}^2$ via Eq.~\eqref{eq:pos_refine}, forming a candidate set $\mathcal{D} = \{(\mathbf{p}_i, c_i)\}_{i=1}^{M}$ with response confidences $c_i$. A greedy count-constrained NMS (Non-Maximum Suppression) then finalizes the selection: candidates are visited in descending confidence, each accepted candidate suppresses neighbors within a distance $d_{\min}$ set to $0.4$ pixels (cf. Sec.~\ref{subsec:dataset}) to preclude duplicate detections, and the procedure terminates exactly when $\hat{N}$ targets have been retained. The count prior thus replaces the fragile threshold as the stopping rule, and the continuous coordinates ensure that the suppression geometry operates on true sub-pixel positions rather than quantized ones. Informed and continuous, the two ideas of this work, literally meet in this final loop.

\subsection{Learning Objective}
\label{sec:objective}

DISTA-Net++ is trained end to end with an objective that mirrors its design: a reconstruction term anchors the backbone to the physical observation model, a counting term teaches the semantic prior, and an offset term supervises continuous rectification. We describe each term and then their coupling.

\noindent \textbf{Reconstruction Loss.}
The primary term enforces fidelity between the unfolded estimate and the ground-truth source distribution while keeping the learnable transforms physically well-behaved as $\mathcal{L}_{\text{rec}} = \mathcal{L}_{\text{discrepancy}} + \gamma\, \mathcal{L}_{\text{constraint}}$, where
\begin{equation}
    \mathcal{L}_{\text{discrepancy}} = \frac{1}{M N_s} \sum_{i=1}^M
    \big\| \tilde{\mathbf{s}}_i^{(N)} - \mathbf{s}_i \big\|_2^2,
    \label{eq:Loss1}
\end{equation}
and
\begin{equation}
    \mathcal{L}_{\text{constraint}} = \frac{1}{M N_s} \sum_{i=1}^M
    \sum_{k=1}^N \big\| \tilde{\mathcal{F}}^{(k)} \big(\mathcal{F}_d^{(k)}
    (\mathbf{s}_i)\big) - \mathbf{s}_i \big\|_2^2.
    \label{eq:Loss2}
\end{equation}
The discrepancy term is the data-fidelity component of the inverse problem, measuring the squared error between the final-stage estimate and the ground truth on the sub-pixel grid. The constraint term imposes stage-wise approximate invertibility $\tilde{\mathcal{F}}^{(k)} \circ \mathcal{F}_d^{(k)} \approx \mathbf{I}$; without it, the learnable transforms would be free to drift into arbitrary mappings, severing the correspondence between network stages and optimization iterations that gives unfolding its interpretability. Since this constraint is a soft structural regularizer rather than a competing objective, we set $\gamma = 0.01$.

\noindent \textbf{Counting Loss.}
The reconstruction residual alone cannot teach the count prior, for it is nearly indifferent to count error: splitting one source into two, or fusing two into one, perturbs the pixel-wise residual only marginally. We therefore supervise the counting branch directly through $K$-way cross-entropy,
\begin{equation}
    \mathcal{L}_{\text{cnt}} = -\frac{1}{M} \sum_{i=1}^{M}
    \log\left( \frac{\exp(z_{i, y_i})}{\sum_{j=0}^{K-1}
    \exp(z_{i, j})} \right),
    \label{eq:loss_cnt}
\end{equation}
where $K$ is the maximum detectable count, $z_{i,j}$ the logit of the $i$-th sample for counting class $j$, and $y_i$ the ground-truth count. This loss penalizes exactly the density-inconsistent solutions that the pixel-wise residual cannot distinguish, shrinking the feasible set of the ill-posed problem toward reconstructions whose sparsity pattern matches the true count. It is the mechanism through which blindness is trained away.

\noindent \textbf{Offset Loss.}
Supervising the offset field poses its own difficulty: a naive dense regression is dominated by the vast background where offsets are undefined, drowning the gradients of the few informative locations. We therefore adopt a mask-driven normalized $\ell_1$ loss,
\begin{equation}
    \mathcal{L}_{\text{off}} = \frac{\sum_{i=1}^{M} \big\| \mathbf{W}_i
    \odot (\mathbf{O}_{i}^{\text{pred}} - \mathbf{O}_{i}^{\text{gt}})
    \big\|_1}{\sum_{i=1}^{M} \| \mathbf{W}_i \|_1},
    \label{eq:loss_off}
\end{equation}
where $\mathbf{W}_i$ is a binary mask activated only at target-occupied cells and the denominator renders the supervision invariant to the per-image target count. Functionally, this loss completes the decoupling promised in Sec.~\ref{subsec:coor}: instead of purchasing precision through ever finer discretization, the network learns the residual displacement directly in the continuous domain, while the masked gradients simultaneously sharpen the backbone's responses at true target locations.

\noindent \textbf{Joint Optimization.}
The total objective combines the three terms as $\mathcal{L}_{\text{total}} = \mathcal{L}_{\text{rec}} + \lambda_1 \mathcal{L}_{\text{cnt}} + \lambda_2 \mathcal{L}_{\text{off}}$, where $\lambda_1$ and $\lambda_2$ control the strength of the semantic and positional regularization, with sensitivity analyzed in Sec.~\ref{sec:ablation}. We stress that the three objectives are not independent tasks sharing a backbone by convenience, but reciprocally constraining views of one problem. The counting loss propagates count-aware gradients into the unfolding stages, suppressing spurious activations, while a cleaner reconstruction in turn yields more discriminative features for count estimation. The offset loss backpropagates fine-grained positional gradients that sharpen the reconstruction peaks, while an accurate grid-level reconstruction supplies reliable anchors for offset regression. This closed loop steers the network toward solutions that are at once physically faithful, semantically consistent, and spatially precise, realizing the informed and continuous paradigm at the level of optimization itself.

\section{Experiments}
\label{sec:experiments}

Our experiments do more than rank DISTA-Net++ against prior art; they are structured to systematically verify the theoretical claims on which this work rests. Five questions organize this evaluation. First, does the shift to an \textit{informed, continuous} formulation decisively shatter both the geometric quantization floor and the semantic count ambiguity that bottleneck existing methods (Sec.~\ref{subsec:sota})? Second, does this advantage hold under stress conditions like varying inter-target distances, dense target clusters, and physical sensor noise (Sec.~\ref{subsec:robustness})? Third, does continuous coordinate rectification decouple localization precision from grid resolution, thereby averting the computational penalty of grid densification (Sec.~\ref{subsec:grid})? Fourth, are the \textit{informed} and \textit{continuous} components portable paradigms that can seamlessly upgrade architecturally disparate backbones (Sec.~\ref{subsec:paradigm_shift})? Fifth, how does each architectural module contribute to the overall performance, and are the internal designs and hyperparameters of the proposed mechanisms robust and well-founded (Sec.~\ref{sec:ablation})?

\subsection{Experimental Settings}
\label{subsec:setting}

\noindent \textbf{Training.} Using CSIST-100K images as input, we apply the sub-pixel division scheme of Sec.~\ref{subsec:proximal} with a sampling ratio $c$ to generate the high-resolution ground truth for the reconstruction backbone: for each target $(x_i, y_i, g_i)$, the intensity $g_i$ is assigned to the cell at $\left(c \cdot x_i + \frac{c - 1}{2},\, c \cdot y_i + \frac{c - 1}{2}\right)$, with all remaining cells set to zero. The ratio $c$ is chosen such that each target occupies a distinct cell, which suffices for separation. As illustrated in Fig.~\ref{fig:subpixels}, the residual displacement between a cell center and the true target position is what the grid cannot express; its signed counterpart serves as the ground truth for the offset head of CCR.

\noindent \textbf{Testing.} At inference, we (1) gather candidate targets on the reconstructed grid with an intensity threshold of 50; (2) project grid coordinates back to the original $11 \times 11$ space via $\left(\frac{x_i-\left\lfloor\frac{c-1}{2}\right\rfloor}{c}, \frac{y_i-\left\lfloor\frac{c-1}{2}\right\rfloor}{c}\right)$; (3) rectify the candidates to continuous coordinates via Eq.~\eqref{eq:pos_refine}; and (4) finalize the selection with greedy count-constrained NMS. The resulting coordinates are matched against ground-truth positions for CSO-mAP, while PSNR and CSO-SSIM are computed between predicted and ground-truth high-resolution maps.

\noindent \textbf{Metrics.} We adopt the metric suite established in Sec.~\ref{sec:ecosystem}, whose components deliberately probe complementary facets of the task: CSO-mAP evaluates the joint quality of separation and sub-pixel localization across distance thresholds; C-ACC isolates exact count estimation at the image level, where a single miscount fails the sample; TP-PRMSE isolates the pure coordinate error of successfully matched targets, and is therefore the direct probe of the quantization floor; PSNR and CSO-SSIM quantify global radiometric fidelity. This decoupled design allows us to attribute each gain to its architectural cause rather than reporting an entangled aggregate.

\noindent \textbf{Implementation.} All experiments run on NVIDIA RTX 3090 GPUs. We use Adam with a learning rate of $10^{-4}$ and gradient clipping ($\text{max\_norm}=1.0$, $\text{norm\_type}=2$), training for 250 epochs with batch size 64 and selecting the checkpoint with the highest CSO-mAP. Following our preliminary work~\cite{han2025dista}, we set $c=3$, $N=6$, $\alpha=0.7$, and $\gamma=0.01$. The two newly introduced loss weights are set to $\lambda_1=250$ and $\lambda_2=300$, with sensitivity analyzed in Sec.~\ref{sec:ablation}.

\begin{figure}[htbp]
    \centering
    \includegraphics[width=\linewidth]{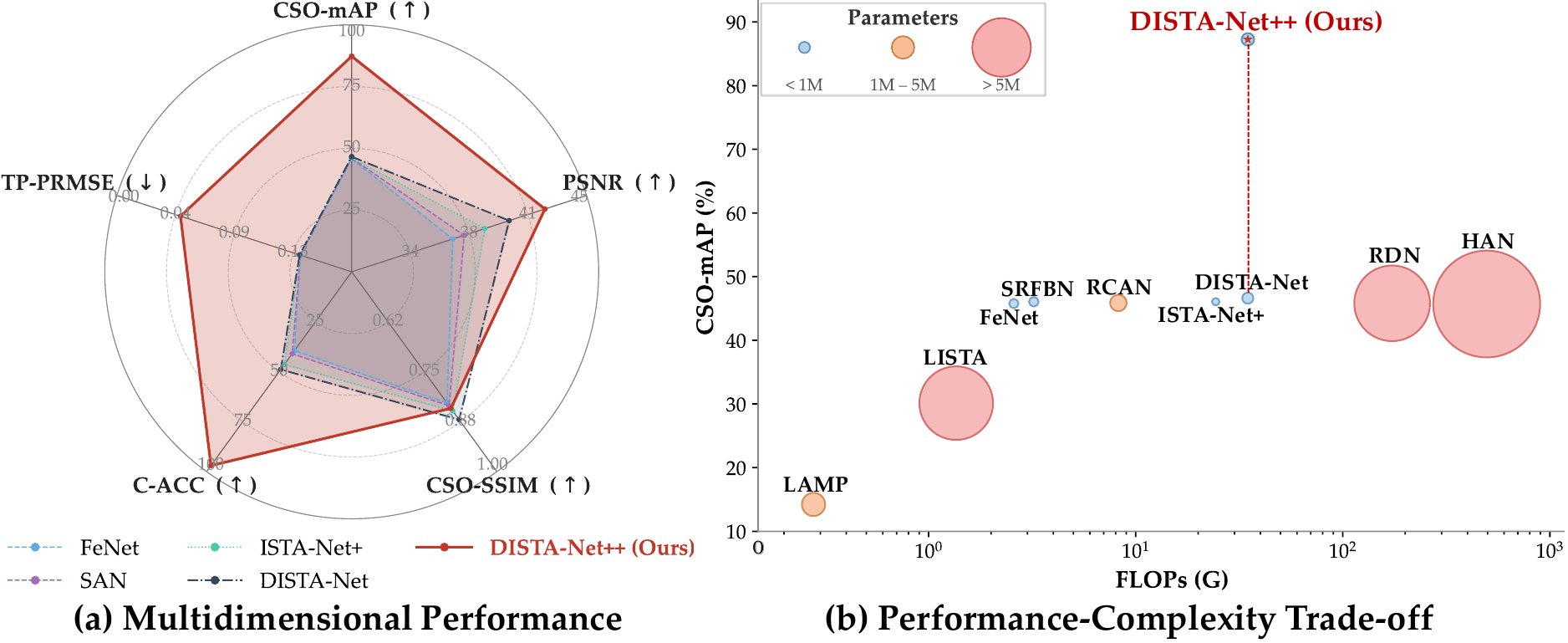} 
    \caption{\textbf{Comprehensive performance and efficiency comparison on CSIST-100K}. \textbf{(a) Multidimensional Performance:} A radar chart benchmarking representative frameworks across five core metrics. Note that the TP-PRMSE ($\downarrow$) axis is intentionally inverted (the outer edge represents a lower, superior error) to maintain a visually consistent ``larger area indicates better overall performance'' paradigm. \textbf{(b) Performance-Complexity Trade-off:} Evaluation of CSO-mAP against computational complexity (FLOPs, mapped to a logarithmic x-axis) and network capacity (parameters, represented by bubble area). DISTA-Net++ achieves an optimal balance, dwarfing the performance of heavy  architectures at merely a fraction of their computational budget.}
    \label{fig:radar_bubble}
\end{figure}

\begin{table*}[htbp]
  \renewcommand\arraystretch{1.2}
  \footnotesize
  \centering
  \caption{Comparison with SOTA methods on the CSIST-100K dataset.}
  \vspace{3pt}
  \setlength{\tabcolsep}{4.pt}
  \begin{tabular}{l|c|c|c|ccccc|c|c|c|c}
  \Xhline{0.8pt}
  \multirow{2}{*}{Method}   & \multirow{2}{*}{Params$\downarrow$} & \multirow{2}{*}{FLOPs$\downarrow$} & \multicolumn{6}{c|}{CSO-mAP}  & \multirow{2}{*}{PSNR$\uparrow$} & \multirow{2}{*}{CSO-SSIM$\uparrow$} & \multirow{2}{*}{C-ACC$\uparrow$} & \multirow{2}{*}{TP-PRMSE$\downarrow$} \\
   & &  & mAP & AP-05  & AP-10    & AP-15    & AP-20  & AP-25 & & & & \\
  
  \Xhline{0.8pt}
  \multicolumn{13}{l}{\textit{Traditional Optimization}}  \\ \hline
  ISTA~\cite{Daubechies2004ISTA} & - & 398.6M & 6.327 & 0.3910 & 1.722 & 4.549 & 9.186 & 15.79 & 26.87 & 0.0590 & 11.68 & 0.1459 \\
  \hline
  \multicolumn{13}{l}{\textit{Image Super-Resolution}}  \\ \hline
  ACTNet~\cite{zhang2023actnet} & 46.21M & 62.80G & 45.51 & 0.3821 & 7.414 & 40.46 & 83.28 & 96.01 & 36.89 & 0.8275 & 42.36 & 0.1329 \\
  \hline
  CTNet~\cite{wang2021contextual} & 0.4003M & 2.756G & 45.30 & 0.3864 & 7.485 & 40.41 & 82.69 & 95.51 & 35.62 & 0.8076 & 35.14 & 0.1331 \\
  \hline
  DCTLSA~\cite{10215496} & 0.8650M &13.56G& 44.47 & 0.3643 & 7.157 & 39.40 & 81.07 & 94.38 & 35.22 & 0.7989 & 33.37 & 0.1334 \\
  \hline
  EDSR~\cite{lim2017enhanced} & 1.552M & 12.04G & 45.45 & 0.3820 & 7.478 & 40.73 & 83.08 & 95.55 & 36.44 & 0.8190 & 37.92 & 0.1332 \\
  \hline
  EGASR~\cite{qiu2023cross} & 2.897M & 17.73G & 45.47 & 0.3906 & 7.640 & 41.18 & 82.84 & 95.28 & 35.46 & 0.8028 & 35.02 & 0.1332 \\
  \hline
  FeNet~\cite{wang2022fenet} & 0.6825M & 5.289G & 45.98 & 0.4139 & 7.889 & 41.85 & 83.84 & 95.90 & 36.45 & 0.8280 & 39.46 & 0.1328 \\
  \hline
  RCAN~\cite{zhang2018image} & 1.079M & 8.243G & 45.88 & 0.3592 & 7.407 & 41.44 & 84.07 & 96.15 & 36.59 & 0.8290 & 39.26 & 0.1328 \\
  \hline
  RDN~\cite{zhang2018residual} & 22.31M & 173.0G & 45.57 & 0.3715 & 7.383 & 40.99 & 83.25 & 95.87 & 36.19 & 0.8160 & 38.48 & 0.1332 \\
  \hline
  SAN~\cite{dai2019second} & 4.442M & 34.05G & 45.93 & 0.3769 & 7.409 & 41.62 & 83.93 & 96.33 & 37.18 & 0.8349 & 41.02 & 0.1329 \\
  \hline
  SRCNN~\cite{dong2015image} & 0.01939M & 1.345G & 33.54 & 0.3911 & 5.759 & 27.36 & 58.55 & 75.66 & 29.47 & 0.5520 & 2.194 & 0.1369 \\
  \hline
  SRFBN~\cite{li2019feedback} & 0.3727M & 3.217G & 45.82 & 0.4291 & 8.351 & 42.48 & 83.22 & 94.62 & 34.73 & 0.8025 & 29.79 & 0.1332 \\
  \hline
  HAN~\cite{niu2020single} & 64.34M & 495.0G & 45.50 & 0.3598 & 7.062 & 39.94 & 83.56 & 96.60 & 36.60 & 0.8265 & 43.98 & 0.1330 \\
  \hline
  HiT‑SNG~\cite{zhang2024hit} & 0.9517M  & 13.32G & 44.75 & 0.3656 & 7.344 & 39.85 & 81.36 & 94.81 & 35.41 & 0.8066 & 38.23 & 0.1331 \\
  \hline
  
  \multicolumn{13}{l}{\textit{Deep Unfolding}}  \\
  \hline
  ISTA-Net~\cite{CVPR2018ISTANet}  & 0.1711M & 12.77G & 45.54 & 0.3373 & 7.037 & 40.86 & 83.91 & 95.55 & 35.67 & 0.8220 & 31.10 & 0.1329 \\
  \hline
  ISTA-Net+~\cite{CVPR2018ISTANet}  & 0.3370M & 24.33G & 45.70 & 0.3526 & 6.962 & 40.63 & 84.21 & 96.32 & 38.50 & \underline{0.8482} & 46.32 & 0.1327 \\
  \hline
  LAMP~\cite{borgerding2017amp} & 2.126M & 0.2783G & 14.04 & 0.0530 & 1.093 & 7.180 & 21.12 & 40.77 & 27.82 & 0.3228 & 9.655 & 0.1456 \\
  \hline
  LIHT~\cite{blumensath2009iterative} & 21.10M & 1.358G & 10.39 & 0.0565 & 0.9255 & 5.075 & 14.84 & 31.08 & 27.56 & 0.2577 & 13.24 & 0.1464 \\
  \hline
  LISTA~\cite{gregor2010learning} & 21.10M & 1.358G & 30.04 & 0.2455 & 4.226 & 22.25 & 50.97 & 72.49 & 30.13 & 0.5763  & 9.195 & 0.1387 \\
  \hline
  FISTA-Net~\cite{xiang2021fista} & 0.07460M & 18.96G & 45.25 & 0.3813 & 7.208 & 40.08 & 83.06 & 95.52 & 38.48 & 0.8266 & 40.68 & 0.1332 \\
  \hline
  TiLISTA~\cite{gregor2010learning} & 2.126M  & 0.2783G & 16.18 & 0.06690 & 1.334 & 8.689 & 24.85 & 45.94 & 27.77 & 0.3362  & 13.40 & 0.1451 \\
  \hline
  DISTA-Net~\cite{han2025dista}   & 0.4911M & 34.76G   & \underline{46.62}   & 0.3410 & 6.969 &41.72 &\underline{86.50} & \textbf{97.57} & \underline{40.06} & \textbf{0.8699} & \underline{48.87} & \underline{0.1323} \\
  \hline
 \rowcolor[rgb]{0.93,0.95,1.0} \textbf{DISTA-Net++ (Ours)} & 0.6210M & 34.79G & \textbf{87.47} & \textbf{65.62} & \textbf{85.65} & \textbf{92.69} & \textbf{95.87} & \underline{97.55} & \textbf{42.78} & 0.8416 & \textbf{97.14} & \textbf{0.04602} \\
  \hline
  \end{tabular}
  \label{tab:msar1}
  \vspace{-1\baselineskip}
\end{table*}

\subsection{Comparison with State-of-the-Art Methods}
\label{subsec:sota}

We benchmark DISTA-Net++ against 22 methods spanning three paradigms: traditional optimization (e.g., ISTA), image super-resolution (e.g., RCAN, HAN, SRFBN), and deep unfolding (e.g., ISTA-Net+, FISTA-Net, DISTA-Net). Results on CSIST-100K are summarized in Table~\ref{tab:msar1} and Fig.~\ref{fig:radar_bubble}(a).

\noindent\textbf{The Quantization Floor Is Real, and It Is Broken.}
Table~\ref{tab:msar1} separates DISTA-Net++ from all baselines not by degree but by kind, and the separation is sharpest exactly where our analysis predicts. Under the stringent thresholds AP-05 to AP-15, every existing method, regardless of paradigm, collapses: no baseline exceeds 10\% at AP-10, with the best (SRFBN) reaching only 8.31\%. This is not an implementation deficiency. At $c=3$, any prediction locked to a cell center carries a worst-case displacement of $\sqrt{2}/(2c) \approx 0.236$ pixels, which categorically precludes matching under tight distance thresholds, however capable the reconstruction network may be. The TP-PRMSE column renders this floor visible with striking clarity: all 20 learning-based baselines, despite differing in architecture by three orders of magnitude in FLOPs, cluster within the razor-thin band of 0.1323 to 0.1334, essentially the statistical error of uniform quantization on a one-third-pixel lattice. Twenty independent architectures converging to the same constant is the empirical signature of a structural bound, not a capacity bound.

DISTA-Net++ breaks through this band decisively, reaching 65.62\% at AP-05, 85.65\% at AP-10, and reducing TP-PRMSE to 0.0461, roughly one third of the shared floor. The overall CSO-mAP rises from 46.62\% (DISTA-Net) to 87.47\%, an absolute gain of 40.85 points. The profile of this gain across thresholds is itself diagnostic. At the loosest AP-25, where the tolerance absorbs quantization error, DISTA-Net++ merely ties its predecessor (97.55\% vs.\ 97.57\%); as the threshold tightens, the margin widens monotonically, exploding to over 65 points at AP-05. This is precisely the signature of genuine off-grid localization, as opposed to better on-grid separation, which would lift results under all thresholds uniformly.

\noindent\textbf{Attributing the Gain to Its Two Sources.}
The decoupled metrics trace the improvement to the two ideas of this work. C-ACC isolates the semantic question: existing methods, reasoning blindly from local intensity patterns, resolve the exact count in under half of the scenes (48.87\% for the best baseline), whereas the Count-Guided Prior lifts DISTA-Net++ to 97.14\%. TP-PRMSE isolates the geometric question, and its drop below the quantization band confirms that CCR performs refinement that no grid-locked formulation can attain. That the two metrics improve simultaneously, and that PSNR also reaches its highest value (42.78~dB, which in this task demands correct positions, counts, and faithful intensities), indicates that neither gain is purchased at the expense of the other.

\noindent\textbf{Efficiency.}
As Fig.~\ref{fig:radar_bubble}(b) shows, these dramatic gains are achieved at minimal computational expense, establishing a new Pareto frontier for the field. DISTA-Net++ requires only 0.627M parameters and 34.789G FLOPs—a fraction of the budget demanded by heavy super-resolution networks such as HAN (495.0G) and RDN (173.0G), and only marginally more than the preliminary DISTA-Net itself. This stark contrast confirms our central thesis: it is the fundamental paradigm upgrade from blind/discrete to informed/continuous, rather than the mere injection of network capacity, that accounts for the empirical breakthrough.

\begin{figure*}[htbp]
    \centering
    \includegraphics[width=\textwidth]{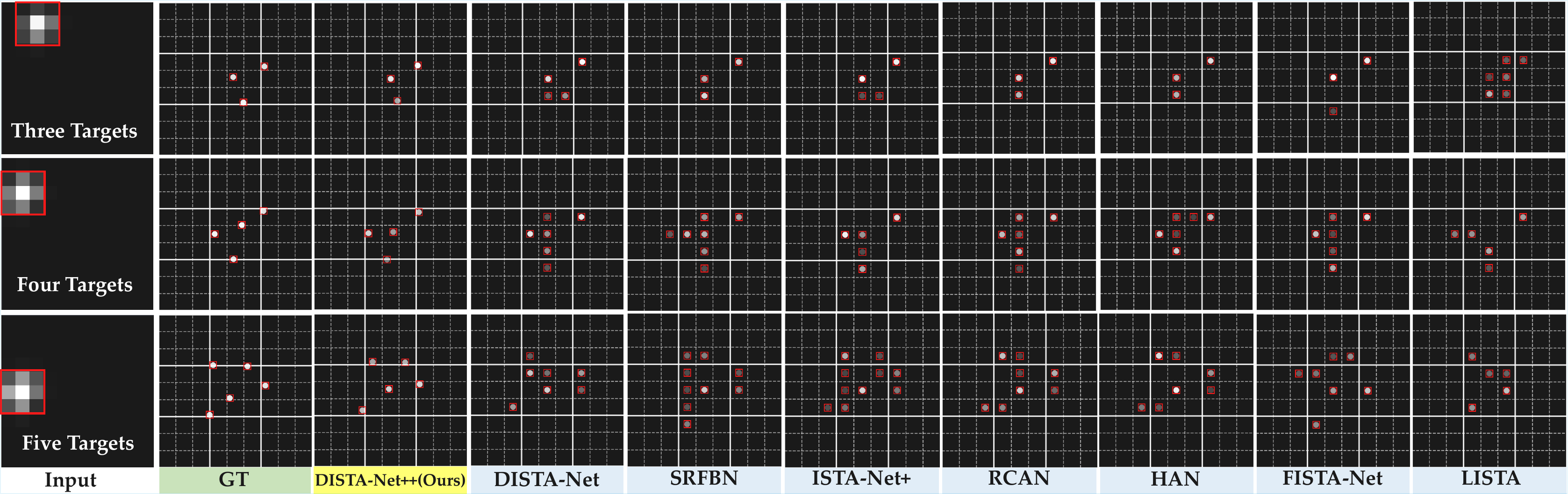}
    \caption{\textbf{Visual comparison of 3$\times$ sub-pixel division unmixing for scenes containing different numbers of closely-spaced infrared small targets}. Each row corresponds to a target count scenario, where the first column shows the raw input with an overlapping diffraction spot highlighted by a red box, and subsequent columns present the unmixing results of different methods for this highlighted region. Unmixing results are shown cropped and enlarged for clarity.}
    \label{fig:compare}
\end{figure*}

\noindent\textbf{Visual Comparison.}
Fig.~\ref{fig:compare} corroborates the quantitative story on challenging dense scenes. The predictions of SRFBN, ISTA-Net+, RCAN, and FISTA-Net are visibly grid-locked, and this forced discretization triggers a cascade of coupled failures: target energy leaks into adjacent cells (degrading PSNR), spurious responses arise as false alarms (failing C-ACC criterion), and every retained prediction inherits an irreducible offset (failing the tight CSO-mAP thresholds). Even DISTA-Net, the strongest unfolding baseline, misses targets and drifts off-center under severe overlap, most evidently in the five-target scenario. DISTA-Net++ instead places compact radiation peaks freely off the grid, aligned with the true sub-pixel coordinates, without energy dilution or spurious artifacts.

\begin{figure*}[t]
    \centering
    \includegraphics[width=\textwidth]{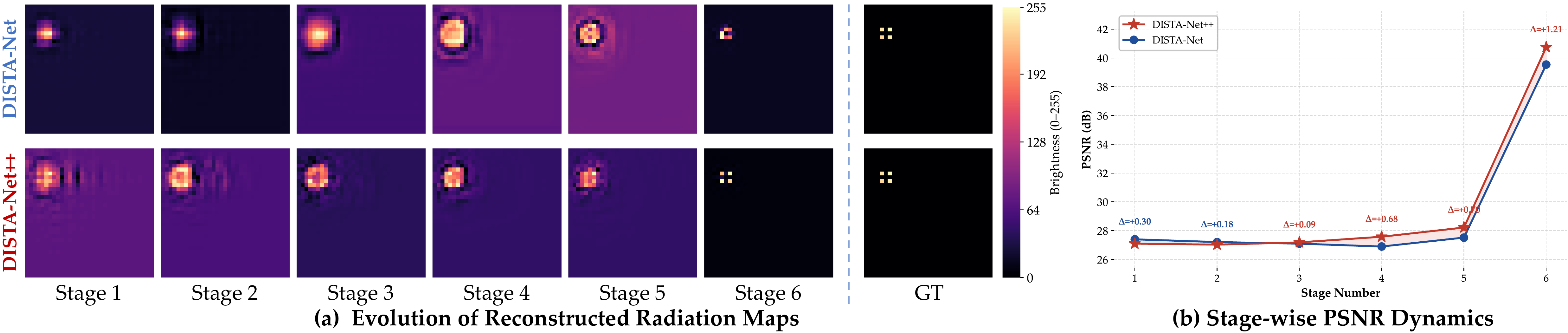}
  \caption{\textbf{Interpretability through stage-wise evolutionary trajectories.}
  \textbf{(a) Evolution of Reconstructed Radiation Maps.} Tracking the intermediate representations from Stage 1 to Stage 6 for a highly challenging dense 4-target cluster (Ground Truth shown on the right). While the baseline DISTA-Net merges the cluster into a diffusive blob and ultimately fails to recover the source count, DISTA-Net++ actively suppresses count ambiguity, rapidly converging to four compact, off-grid spatial peaks.
  \textbf{(b) Stage-wise PSNR Dynamics.} PSNR evolution under identical unfolding depths. The terminal spike reflects the final structural projection from continuous feature manifolds to the discrete radiation map, where DISTA-Net++ decisively outperforms the baseline by expanding the fidelity margin ($\Delta=+1.21$ dB).}
  \label{fig:stage_and_psnr}
\end{figure*}

\noindent\textbf{Stage-wise Evolution.}
A key virtue of deep unfolding is its interpretable stage-wise trajectory, which enables diagnosis beyond endpoint metrics.
Fig.~\ref{fig:stage_and_psnr} illustrates a typical failure mode of \emph{blind, discrete} refinement in DISTA-Net: early stages can look plausible, yet later updates drift toward grid-locked and diffusive patterns, where energy leakage and slight mis-centering persist rather than being corrected (Fig.~\ref{fig:stage_and_psnr}(a)).
DISTA-Net++ makes the trajectory \emph{informed} and \emph{continuous}: the Count-Guided Prior reduces count ambiguity at the scene level, while CCR breaks the grid-lock via off-grid rectification.
As a result, compact peaks emerge and remain stable across stages with less diffusion and error accumulation (Fig.~\ref{fig:stage_and_psnr}(a)), and PSNR stays consistently higher with a clearer late-stage margin (Fig.~\ref{fig:stage_and_psnr}(b)).

\subsection{Robustness under Stress Conditions}
\label{subsec:robustness}

As detailed in Sec.~\ref{subsec:dataset}, CSIST-100K provides an idealized, strictly controlled baseline (noise-free mixtures of 1--5 closely spaced targets). To better reflect real IRST operational extremes, we stress-test the models using the same simulation pipeline along three harder dimensions: inter-target distance, target count, and background noise.

\begin{figure}[htbp]
    \centering
    \includegraphics[width=1.0\columnwidth]{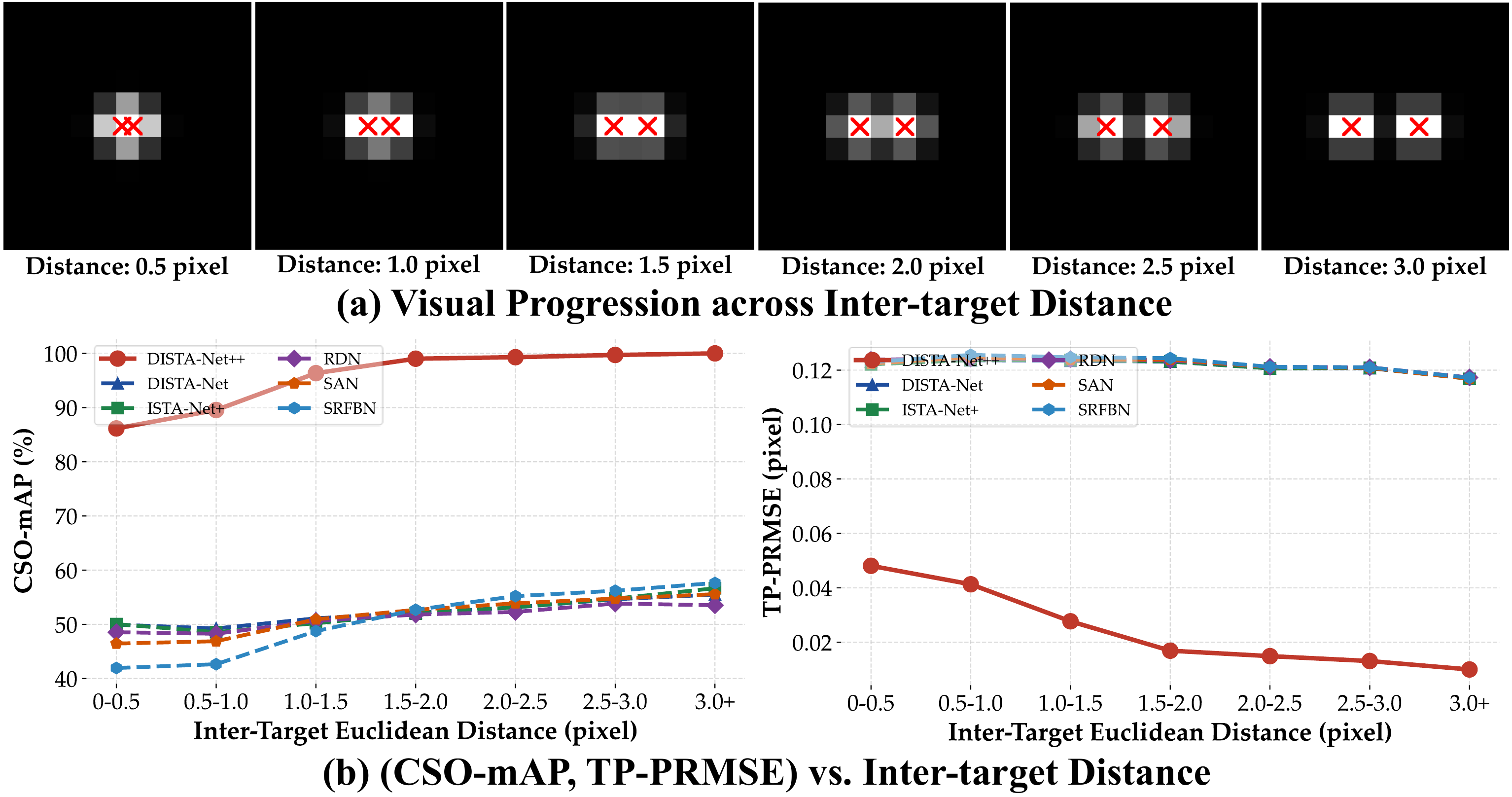}
    \caption{\textbf{Robustness across extended inter-target distances.} \textbf{(a) Visual Progression across Inter-target Distances.} Evolution from severely mixed ($0.5$ pixel) to fully isolated ($3.0$ pixels). \textbf{(b) (CSO-mAP, TP-PRMSE) vs.\ Inter-target Distance.} While the localization error of grid-based baselines remains clamped to their theoretical quantization floor regardless of inter-target distance, DISTA-Net++ utilizes continuous regression to drive sub-pixel error toward zero as spatial overlap relaxes.}
    \label{fig:distance}
\end{figure}

\noindent\textbf{Extended Inter-target Distance.}
A truly robust unmixing framework must accurately localize targets not only when they heavily overlap but also as they separate into isolated sources. Fig.~\ref{fig:distance} evaluates performance as the inter-target Euclidean distance extends from 0 to over 3.0 pixels. The results expose a structural rigidity in prior art: even as targets become visually distinct (Fig.~\ref{fig:distance}(a)), the localization error (TP-PRMSE) of all baselines remains uniformly stalled above 0.12 pixels. This confirms that their precision is permanently bottlenecked by the discrete grid formulation, independent of target density. In contrast, DISTA-Net++ decouples precision from grid resolution. 
As inter-target distance increases, its TP-PRMSE continuously approaches zero and CSO-mAP converges to near-perfection, confirming that the Continuous Coordinate Rectification (CCR) behaves as a robust off-grid solver across varying spatial distributions.

\begin{figure}[htbp]
    \centering
    \includegraphics[width=1.0\columnwidth]{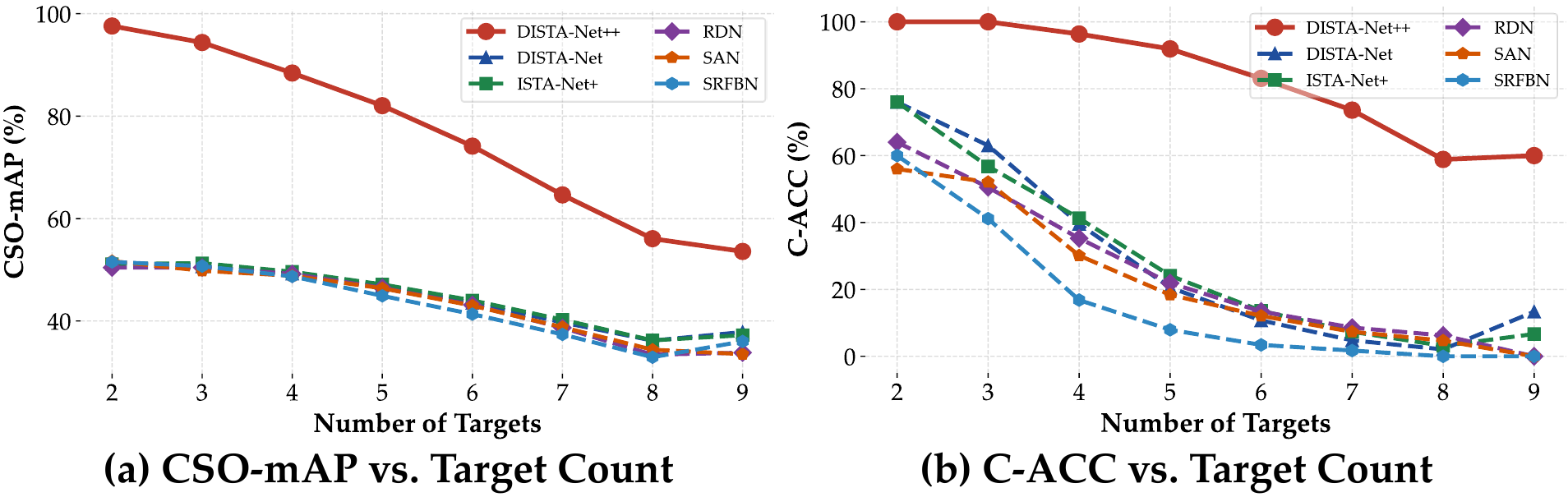}
    \caption{\textbf{Robustness under extended target count.} Evaluation as cluster density scales across the expanded range (plotted from 2 to 9 targets). \textbf{(a) CSO-mAP vs.\ Target Count.} The trajectory of overall sub-pixel detection performance. \textbf{(b) C-ACC vs.\ Target Count.} The trajectory of semantic count estimation. Even when trained on datasets containing these extremely dense configurations, baselines suffer systemic semantic collapse, whereas DISTA-Net++ sustains a commanding accuracy lead.}
    \label{fig:number}
\end{figure}

\noindent\textbf{Extended Target Count.}
The ultimate test of semantic unmixing is its robustness to dense, saturated clusters. Fig.~\ref{fig:number} evaluates the models on the expanded dataset, tracking performance as the target count scales up. Crucially, since all models are trained on configurations containing up to 10 targets, the performance drops observed here reveal fundamental limits in their formulations rather than a lack of exposure to dense scenes. As the plotted target count reaches 6 and beyond, prior art experiences a systemic collapse: the Count Accuracy (C-ACC) of the baselines plummets to near $0\%$ at the extremes (8 to 9 targets). This proves that blind, grid-based unmixing physically cannot disentangle severe overlaps, regardless of how much dense training data it is fed. DISTA-Net++, fortified by its explicit semantic guidance, successfully maintains its disentanglement capability, retaining approximately $60\%$ C-ACC even at the extreme density of 9 targets. This indicates that the informed continuous formulation is intrinsically resilient to spatial crowding, decisively overcoming the structural bottlenecks of legacy paradigms.

\begin{figure}[htbp]
    \centering
    \includegraphics[width=1.0\columnwidth]{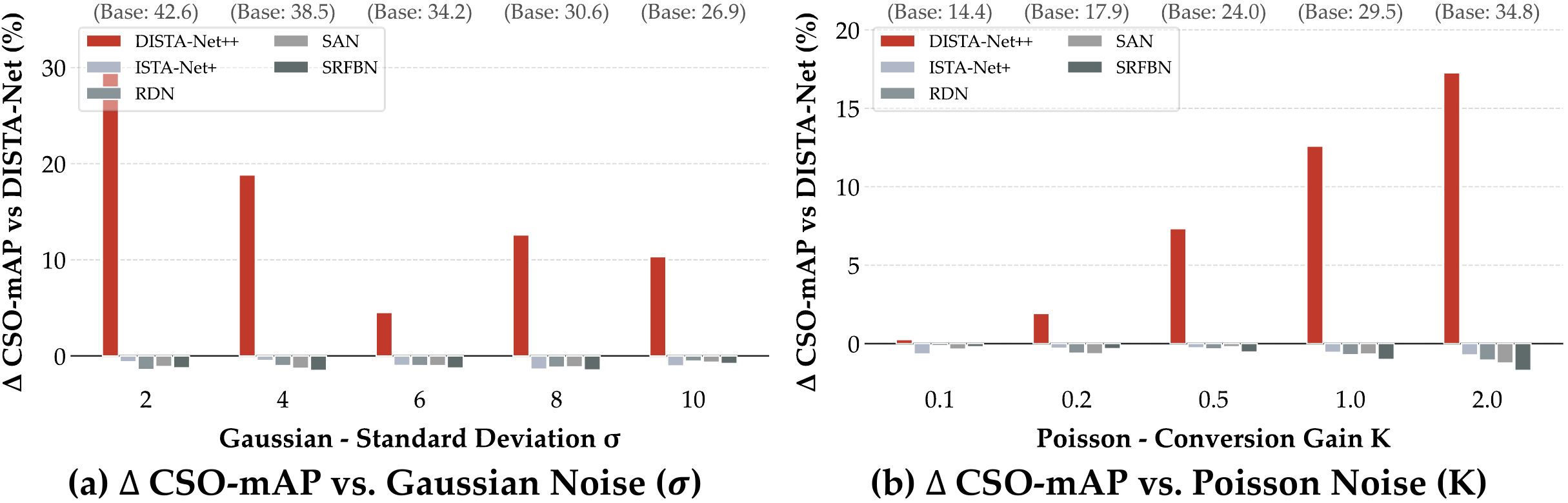}
    \caption{\textbf{Robustness against physics-based sensor noise.} Relative performance margin ($\Delta$ CSO-mAP) compared to the DISTA-Net baseline under physically faithful degradation models. \textbf{(a) $\Delta$ CSO-mAP vs.\ Gaussian Standard Deviation ($\sigma$).} Performance under varying thermal noise standard deviations. \textbf{(b) $\Delta$ CSO-mAP vs.\ Poisson Conversion Gain ($K$).} Performance under varying photon conversion gains. DISTA-Net++ demonstrates remarkable resilience, maintaining substantial positive margins even as severe physical noise uniformly collapses the baseline performances.}
    \label{fig:noise}
\end{figure}

\noindent\textbf{Physics-Based Sensor Noise.}
While the standard benchmark assumes ideal imaging, practical infrared detectors are fundamentally bounded by the physics of photon sensing and electronic readout. To rigorously simulate this, we evaluate performance under two degradation paradigms. The first is Additive White Gaussian Noise (AWGN), simulating basic thermal fluctuations. The second, crucially, is a physically faithful Poisson-Gaussian detector model that sequentially simulates the full imaging pipeline: background flux coupling, photon shot noise in the electron domain, electronic read noise, conversion gain modulation, and final ADC quantization.

Fig.~\ref{fig:noise} demonstrates the relative performance resilience when these unobserved degradations are injected. As noise intensifies, all blind reconstruction baselines degrade sharply and uniformly, as their reliance on purely local intensity mapping renders them highly vulnerable to signal fluctuations. DISTA-Net++, however, degrades far more gracefully, registering massive relative gains over the baselines. This confirms that the global Count acts as a strong structural regularizer; while severe physical noise inevitably corrupts per-pixel fidelity, the semantic abstraction of target count remains remarkably stable, preventing the network from hallucinating false targets from shot noise or read noise spikes.

\begin{table}[htbp]
\setlength{\abovecaptionskip}{0pt} 
\setlength{\belowcaptionskip}{0pt} 
  \caption{Comparison of methods across different sampling grids on the CSIST-100K dataset.}
  \renewcommand\arraystretch{1.2}
  \footnotesize
  \centering
  
  \vspace{-2pt}
  \resizebox{0.99\linewidth}{!}{
  \setlength{\tabcolsep}{6.pt}
  \begin{tabular}{l|cc|cccc}
  \Xhline{1pt}
  \multirow{2}{*}{Method}    & \multirow{2}{*}{Params $\downarrow$} & \multirow{2}{*}{FLOPs $\downarrow$} & \multicolumn{4}{c}{Performance Metrics}  \\
   & &  & CSO-mAP $\uparrow$ & PSNR $\uparrow$ & C-ACC $\uparrow$ & TP-PRMSE $\downarrow$ \\
   
  \Xhline{1pt}
  \multicolumn{7}{l}{\textit{c=5}}  \\
  \hline
  ISTA-Net ~\cite{CVPR2018ISTANet} & 0.1711M  & 39.54G   & 65.54 & 37.96 & 26.80 & 0.08874 \\
  \hline
  ISTA-Net+ ~\cite{CVPR2018ISTANet}& 0.2247M & 48.16G & 68.18 & 40.57 & \underline{35.96} & 0.08761 \\
  \hline
  CFGN ~\cite{dai2023cfgn} & 0.5376M & 4.122G & 68.25 & 39.23 & 34.62 & 0.08772 \\
  \hline
  DISTA-Net ~\cite{han2025dista} & 0.4911M & 99.51G & \underline{69.51} & \underline{41.08} & 35.68 & \underline{0.08727} \\
  \hline
  \rowcolor[rgb]{0.93,0.95,1.0} \textbf{DISTA-Net++ (Ours)} & 0.6273M & 99.54G & \textbf{83.69} & \textbf{42.65} & \textbf{92.62} & \textbf{0.05021} \\
  \hline
  \multicolumn{7}{l}{\textit{c=7}}  \\
  \hline
  ISTA-Net ~\cite{CVPR2018ISTANet} & 0.1711M & 89.51G & 64.27 & 38.71 & 26.95 & 0.07231 \\
  \hline
  ISTA-Net+ ~\cite{CVPR2018ISTANet} & 0.2247M & 103.1G & 70.55 & 40.96 & 32.30 & 0.07219 \\
  \hline
  CFGN ~\cite{dai2023cfgn} & 0.5480M & 4.202G & 68.31 & 39.35 & 33.88 & \underline{0.06901} \\
  \hline
  DISTA-Net ~\cite{han2025dista} & 0.3274M & 137.3G & \underline{71.32} & \underline{41.30} & \underline{34.18} & 0.07064  \\
  \hline
  \rowcolor[rgb]{0.93,0.95,1.0} \textbf{DISTA-Net++ (Ours)} & 0.4511M & 137.3G & \textbf{82.70} & \textbf{43.08} & \textbf{85.39} & \textbf{0.04823} \\
  \hline
  \end{tabular}
  }
  \label{tab:sampling_grid}
  \vspace{-0.5\baselineskip}
\end{table}

\subsection{Does Continuity Decouple Precision from the Grid?}
\label{subsec:grid}

A central claim of this work is that CCR severs the historical coupling between localization precision and grid resolution. This claim admits a sharp test: vary the sampling ratio $c$ and observe whether performance follows the grid. Table~\ref{tab:sampling_grid} reports all competitive methods at $c=5$ and $c=7$.

For discrete paradigms, densification is the only route to precision, and the results confirm both the coupling and its punishing economics. DISTA-Net improves from 69.51\% to 71.32\% CSO-mAP as $c$ grows from 5 to 7, a gain of under two points, while its FLOPs roughly double from 99.5G to 203.7G. Precision is being purchased at quadratically inflating cost, as the dictionary-growth analysis in Sec.~\ref{subsec:coor} predicts.

DISTA-Net++ behaves qualitatively differently, and the difference is the finding. At $c=5$ it already reaches 83.69\% CSO-mAP and 92.62\% C-ACC, exceeding the best baseline by 14.18 points. Increasing the ratio to $c=7$ then \textit{degrades} its performance (82.70\% CSO-mAP, 85.39\% C-ACC) while doubling computation. This inversion is not an anomaly but a consequence of the theory: once the offset head resolves positions in the continuous domain, a finer grid contributes no additional precision; it only inflates the dimension of the sparse solution space, drives adjacent steering vectors toward collinearity, and thereby destabilizes the reconstruction it was meant to sharpen. The practical corollary is striking. Operating at the economical $c=3$ division (34.8G FLOPs), DISTA-Net++ outperforms every baseline running at $c=7$ (up to 203.7G FLOPs) by over 16 points of CSO-mAP, at one sixth of their computation. For grid-based unmixing, densification was the price of precision; DISTA-Net++ declines to pay it.

\subsection{Paradigm Shift Verification: Informed and Continuous as Portable Paradigms}
\label{subsec:paradigm_shift}

A key assertion of this work is that the two proposed paradigms—the \textit{informed} paradigm, realized by the Count-Guided Prior (CGP) together with the Count-Constrained Inference (CCI), and the \textit{continuous} paradigm, realized by the Continuous Coordinate Rectification (CCR)—represent a fundamental shift from blind, discrete separation to informed, continuous unmixing, rather than ad-hoc tweaks tailored exclusively to DISTA-Net++. To rigorously verify this claim, we evaluate the plug-and-play generality of both paradigms by migrating them into representative baselines across Deep Unfolding (e.g., FISTA-Net, ISTA-Net+) and Image Super-Resolution (e.g., RCAN, HAN, SAN, RDN) frameworks.

% ------------------ Continuous 范式通用性扩展表 ------------------
\begin{table}[t]
  \centering
  \caption{Plug-and-play generality of the \textit{continuous} paradigm (CCR).}
  \label{tab:paradigm_shift_continuous}
  \renewcommand\arraystretch{1.15}
  \setlength{\tabcolsep}{6pt}
  \scriptsize
  \vspace{2pt}
  \begin{tabular}{l|cccc}
    \Xhline{0.8pt}
    Model Baseline & CSO‑mAP $\uparrow$ & PSNR$\uparrow$ & C‑ACC $\uparrow$ & TP‑PRMSE$\downarrow$ \\
    \hline
    \multicolumn{5}{l}{\textit{Deep Unfolding}} \\
    \hline
    FISTA‑Net~\cite{xiang2021fista} & 44.66 & 38.48 & 40.68 & 0.1331 \\
    \rowcolor[rgb]{0.93,0.95,1.0} \textbf{\quad + CCR} & \textbf{79.87} & \textbf{38.64} & \textbf{42.39} & \textbf{0.06362} \\
    \hline
    ISTA‑Net+~\cite{CVPR2018ISTANet} & 46.06 & 38.50 & 46.32 & 0.1327 \\
    \rowcolor[rgb]{0.93,0.95,1.0} \textbf{\quad + CCR} & \textbf{85.88} & \textbf{40.23} & \textbf{57.70} & \textbf{0.05185} \\
    \hline
    DISTA‑Net~\cite{han2025dista} & 46.62 & 40.06 & 48.87 & 0.1323 \\
    \rowcolor[rgb]{0.93,0.95,1.0} \textbf{\quad + CCR} & \textbf{88.90} & \textbf{41.23} & \textbf{59.30} & \textbf{0.04702} \\
    \hline
    \multicolumn{5}{l}{\textit{Image Super‑Resolution}} \\
    \hline
    RCAN~\cite{zhang2018image} & 45.87 & 36.59 & 39.26 & 0.1328 \\
    \rowcolor[rgb]{0.93,0.95,1.0} \textbf{\quad + CCR} & \textbf{77.53} & \textbf{39.09} & \textbf{59.78} & \textbf{0.06469} \\
    \hline
    HAN~\cite{niu2020single} & 45.70 & 36.60 & 43.98 & 0.1330 \\
    \rowcolor[rgb]{0.93,0.95,1.0} \textbf{\quad + CCR} & \textbf{72.08} & \textbf{41.11} & \textbf{77.29} & \textbf{0.06661} \\
    \hline
    SAN~\cite{dai2019second} & 45.95 & 37.18 & 41.02 & 0.1329 \\
    \rowcolor[rgb]{0.93,0.95,1.0} \textbf{\quad + CCR} & \textbf{74.82} & \textbf{42.09} & \textbf{76.48} & \textbf{0.06366} \\
    \Xhline{0.8pt}
  \end{tabular}
\end{table}

\noindent\textbf{Continuous Paradigm.}
As detailed in Table~\ref{tab:paradigm_shift_continuous}, migrating the continuous paradigm to any existing baseline consistently shatters the intrinsic quantization barrier ($\text{TP-PRMSE} \approx 0.133$), causing the localization error to plummet down to $0.0519$--$0.0666$. Correspondingly, CSO-mAP scores experience massive gains across all backbones (e.g., ISTA-Net+ leaps from $46.06$ to $85.88$). This model-agnostic breakthrough confirms that spatial quantization is a structural limitation of grid-locked formulations, and that escaping the lattice is a portable principle rather than an architecture-specific trick.

% ------------------ Informed 范式通用性扩展表 ------------------
\begin{table}[t]
  \centering
  \caption{Plug-and-play generality of the \textit{informed} paradigm (CGM + CCI). }
  \label{tab:paradigm_shift_informed}
  \renewcommand\arraystretch{1.15}
  \setlength{\tabcolsep}{6pt}
  \scriptsize
  \vspace{2pt}
  \begin{tabular}{l|cccc}
    \Xhline{0.8pt}
    Model Baseline & CSO-mAP $\uparrow$ & PSNR$\uparrow$ & C-ACC $\uparrow$ & TP-PRMSE$\downarrow$ \\
    \hline
    \multicolumn{5}{l}{\textit{Deep Unfolding}} \\
    \hline
    ISTA-Net+~\cite{CVPR2018ISTANet} & 46.06 & 38.50 & 46.32 & 0.1327 \\
    \quad + CGM & \textbf{46.57} & 39.48 & 48.80 & 0.1326 \\
    \quad + CCI & 41.38 & 38.36 & 74.84 & \textbf{0.1313} \\
    \rowcolor[rgb]{0.93,0.95,1.0} \textbf{\quad + CGM \& CCI} & 43.75 & \textbf{39.97} & \textbf{85.92} & 0.1320 \\
    \hline
    DISTA-Net~\cite{han2025dista} & 46.62 & 40.06 & 48.87 & 0.1323 \\
    \quad + CGM & \textbf{46.76} & 39.49 & 47.39 & 0.1325 \\
    \quad + CCI & 43.89 & 39.50 & \textbf{87.75} & 0.1322 \\
    \rowcolor[rgb]{0.93,0.95,1.0} \textbf{\quad + CGM \& CCI} & 43.84 & \textbf{40.19} & \textbf{87.75} & \textbf{0.1319} \\
    \hline
    \multicolumn{5}{l}{\textit{Image Super-Resolution}} \\
    \hline
    SAN~\cite{dai2019second} & 45.95 & \textbf{37.18} & 41.02 & 0.1329 \\
    \quad + CGM & \textbf{46.10} & 36.68 & 43.03 & 0.1328 \\
    \quad + CCI & 40.26 & 34.94 & 67.91 & 0.1319 \\
    \rowcolor[rgb]{0.93,0.95,1.0} \textbf{\quad + CGM \& CCI} & 40.80 & 37.02 & \textbf{80.17} & \textbf{0.1317} \\
    \Xhline{0.8pt}
  \end{tabular}
\end{table}

\noindent\textbf{Informed Paradigm.}
Table~\ref{tab:paradigm_shift_informed} examines the semantic axis of the shift, decomposing the count modulation into its two channels before assembling them. The continuous embedding alone (CGM only) sharpens source separation on unfolding backbones (ISTA-Net+: $46.06 \rightarrow 46.57$; DISTA-Net: $46.62 \rightarrow 46.76$) and on SAN ($45.95 \rightarrow 46.10$), as the explicit semantic anchor suppresses diffraction sidelobes in sparse scenes while protecting weak overlapping signals in dense ones. The discrete prediction alone (CCI only) transforms counting reliability, surging C-Acc from $46.32$ to $74.84$ on ISTA-Net+, from $41.02$ to $67.91$ on SAN, and from $38.48$ to $86.38$ on RDN. Assembling both channels into the full informed paradigm compounds these effects: since the count embedding is jointly optimized with the reconstruction, the injection in turn sharpens the counting head itself, lifting C-ACC further to $85.92$ (ISTA-Net+), $80.17$ (SAN), and $87.07$ (RDN), while delivering the best PSNR within nearly every group (e.g., $40.1906$ on DISTA-Net). The accompanying mild recession in CSO-mAP is an expected and diagnostic artifact of the discrete regime: when quantized positions err by up to half a grid cell, enforcing exactly $\hat{N}$ detections trades a few loosely-localized true positives for count fidelity. Crucially, this price is paid to quantization, not to the modulation itself—all informed variants remain pinned at the quantization barrier ($\text{TP-PRMSE} \approx 0.132$), reaffirming that semantic guidance alone cannot compensate for a discrete coordinate system.

\begin{table*}[t]
\setlength{\abovecaptionskip}{0pt} 
\setlength{\belowcaptionskip}{0pt} 
\caption{Ablation studies on different components of DISTA-Net++ on the \textbf{CSIST-100K} dataset. DT and DTG form the adaptive backbone; CGM and CCI realize the \textit{informed} paradigm; CCR realizes the \textit{continuous} paradigm.}
\centering
\renewcommand\arraystretch{1.2}
\setlength{\tabcolsep}{4pt}
\scriptsize{
\begin{tabular}{cIcccccIccIcccccc} 
\Xhline{0.8pt}
Config. & DT & DTG & CGM & CCI & CCR & Params (M) $\downarrow$ & FLOPs (G) $\downarrow$ & CSO‑mAP $\uparrow$ & PSNR $\uparrow$ & C‑ACC $\uparrow$ & TP‑PRMSE $\downarrow$ \\ 
\Xhline{0.8pt}
1 & -- & -- & -- & -- & -- & 0.4493 & 31.93 & 46.31 & 36.84 & 34.28 & 0.1328 \\
2 & \checkmark & -- & -- & -- & -- & 0.4494 & 31.93 & 46.25 & 36.56 & 32.84 & 0.1328 \\
3 & -- & \checkmark & -- & -- & -- & 0.4910 & 34.75 & 46.45 & 38.11 & 40.75 & 0.1326 \\
4 & \checkmark & \checkmark & -- & -- & -- & 0.4911 & 34.76 & 46.62 & 40.06 & 48.87 & 0.1323 \\
5 & \checkmark & \checkmark & \checkmark & -- & -- & 0.5641 & 34.79 & 46.76 & 39.49 & 47.39 & 0.1325 \\
6 & \checkmark & \checkmark & -- & \checkmark & -- & 0.5267 & 34.79 & 43.89 & 39.50 & \underline{87.75} & 0.1322 \\
7 & \checkmark & \checkmark & \checkmark & \checkmark & -- & 0.5641 & 34.79 & 43.84 & 40.19 & \underline{87.75} & 0.1319 \\
8 & -- & -- & -- & -- & \checkmark & 0.5062 & 31.93 & \underline{87.49} & 38.25 & 45.62 & 0.04933 \\
9 & \checkmark & \checkmark & -- & -- & \checkmark & 0.5480 & 34.76 & \textbf{88.90} & \underline{41.23} & 59.30 & \underline{0.04702} \\
\rowcolor[rgb]{0.93,0.95,1.0} 10 & \checkmark & \checkmark & \checkmark & \checkmark & \checkmark & 0.6210 & 34.79 & 87.47 & \textbf{42.78} & \textbf{97.14} & \textbf{0.04602} \\
\hline
\end{tabular}
}
\label{tab:ablation_full}
\vspace{-0.3cm}
\end{table*}

\subsection{Ablation Studies}
\label{sec:ablation}

\noindent\textbf{Component Analysis.}
Table~\ref{tab:ablation_full} dissects DISTA-Net++ into its five components: the Dynamic Transform (DT) and Dynamic Thresholding Generator (DTG) that mainly constitute the backbone, the Count-Guided Modulating (CGM) and Count-Constrained Inference (CCI) that realize the \textit{informed} paradigm, and the Continuous Coordinate Rectification (CCR) that realizes the \textit{continuous} paradigm. Three observations emerge, each mapping onto a claim of Sec.~\ref{sec:DISTA_PLUS}.

\textit{First, input adaptivity earns its place.} Rows 1 to 4 evaluate the dynamic backbone inherited from our preliminary work. DT alone (Row 2) changes little, but its joint integration with DTG (Row 4) lifts PSNR from 36.84 to 40.05 dB and C-ACC by 14.59 points over the static baseline, confirming that a transform and threshold conditioned on the observation resolve overlap patterns that frozen parameters cannot.

\textit{Second, quantization, not capacity, is the localization bottleneck.} Comparing Row 1 with Row 8, attaching CCR to an otherwise entirely static backbone raises CSO-mAP from 46.31\% to 87.49\% and cuts TP-PRMSE by nearly two thirds (0.1328 to 0.0493). That the single largest gain in the entire study arrives without touching reconstruction capacity is the cleanest possible verification of our diagnosis: the ceiling on localization was never the network, it was the lattice the network was asked to speak through.

\textit{Third, the two priors need each other.} Rows 4 to 7 probe count guidance. Count-guided prior raises the strict C-ACC from 48.87\% to 87.75\% (Row 6), yet the same row reveals a seeming paradox: CSO-mAP simultaneously drops from 46.62\% to 43.89\%. The explanation lies in the distance-based matching of CSO-mAP (Eq.~\ref{eq:cso-mAP}). Enforcing an exact global count on top of coordinates that are still quantized forces the suppression of nearby candidates that would have matched under looser thresholds; count consistency is bought with average precision. The paradox dissolves the moment continuous coordinates enter: the full model (Row 10) attains the best C-ACC (97.14\%), PSNR (42.78 dB), and TP-PRMSE (0.0461) simultaneously, with CSO-mAP restored to 87.47\%. This interaction substantiates a central claim of the paper at the level of measurement: informed and continuous are not independent add-ons whose benefits sum, but mutually dependent halves of one formulation. The count modulation can only prune correctly when coordinates are accurate, and continuous localization only translates into scene-level correctness when the candidate set is density-consistent.

\begin{figure}[htbp]
    \centering
    \includegraphics[width=1.0\columnwidth]{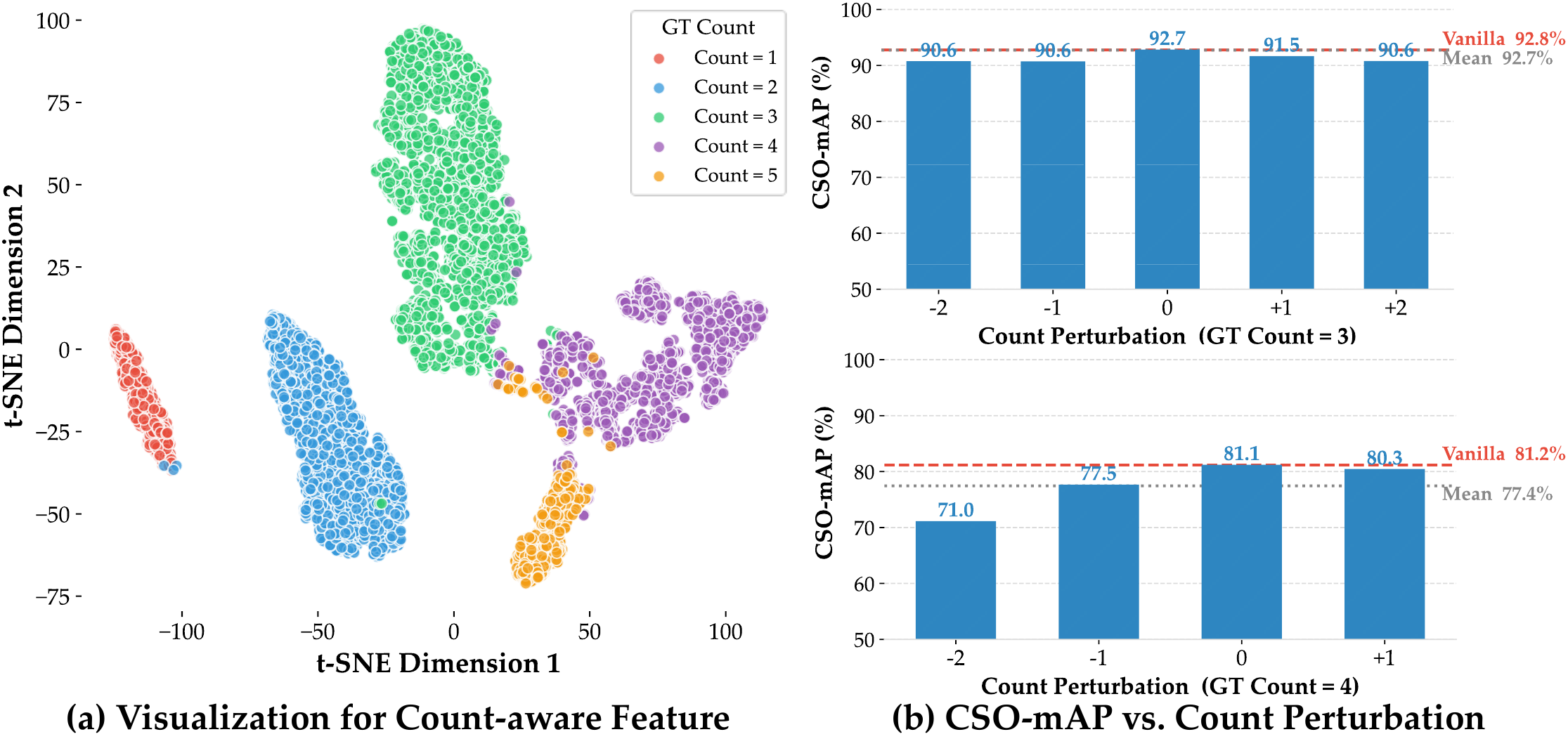}
    \caption{\textbf{Mechanism validation of the Count-Guided Prior.} \textbf{(a) Visualization for Count‑aware Feature.} A t-SNE projection of the count embedding $\mathbf{c}$, demonstrating distinct semantic clustering aligned with the ground-truth target count. \textbf{(b) CSO-mAP vs.\ Count Perturbation.} Performance under count perturbations. Injecting the exact ground-truth embedding (0 perturbation) establishes the theoretical upper bound, which the default predictive model (Vanilla) closely matches. Injecting erroneous embeddings ($\pm 1, \pm 2$) demonstrates graceful degradation, proving the framework's robustness against counting errors.}
    \label{fig:count_exp}
\end{figure}

\noindent\textbf{Mechanism Validation of the Count-guided Prior.}
Fig.~\ref{fig:count_exp} dissects the mechanics of the Count-Guided Prior from three angles: representation, sufficiency, and fault tolerance. The t-SNE projection of the count embedding of counting network in Fig.~\ref{fig:count_exp}(a) forms cohesive, well-separated clusters aligned with the ground-truth count, confirming that the counting network resolves scene-level count without being confused by local intensity overlaps. To test whether this predicted prior suffices, we intercept the forward pass and inject embeddings sampled from pools grouped by ground-truth count, thereby decoupling the prior from the observation: the ground-truth injection (Oracle, $0$ perturbation) proves statistically indistinguishable from the default model (Vanilla)—92.7\% vs.\ 92.8\% CSO-mAP at count 3, and 81.1\% vs.\ 81.2\% at count 4 (Fig.~\ref{fig:count_exp}(b))—indicating that the lightweight counting branch already exhausts the potential of the count-guided formulation. Finally, injecting erroneous embeddings quantifies fault tolerance: even the worst case tested, unmixing a 4-target cluster under a 2-target prior, retains 71.0\% CSO-mAP, still far above all blind baselines (Sec.~\ref{sec:ablation}); and since the contiguous cluster boundaries in Fig.~\ref{fig:count_exp}(a) imply that realistic counting errors are almost exclusively $\pm 1$ deviations, the practical penalty is marginal (e.g., 91.5\% for $+1$ on a 3-target cluster). This resilience is by design: CAM (Eqs.~\ref{eq:cam} and~\ref{eq:theta_count}) treats the prior as a soft modulator of adaptive thresholding rather than a hard switch, so continuous visual evidence always retains a veto over an imperfect count prior.

\begin{table}[t]
  \centering
  \caption{Design analysis of the CCR on CSIST-100K.}
  \label{tab:ccr_design}
  \renewcommand\arraystretch{1.15}
  \setlength{\tabcolsep}{6pt}
  \scriptsize
  \vspace{2pt}
  \begin{tabular}{l|cccc}
    \Xhline{0.8pt}
    Variant & CSO-mAP$\uparrow$ & PSNR$\uparrow$ & C-ACC$\uparrow$ & TP-PRMSE$\downarrow$ \\
    \hline
    w/o $\mathbf{F}_0$ (final only) & 81.46 & 38.28 & 84.64 & 0.04841 \\
    w/o $\tanh$ (linear head)               & 87.07 & 42.47 & 96.92 & \textbf{0.04567} \\
    \rowcolor[rgb]{0.93,0.95,1.0} \textbf{DISTA-Net++ (Ours)} & \textbf{87.47} & \textbf{42.78} & \textbf{97.14} & 0.04602 \\
    \Xhline{0.8pt}
  \end{tabular}
\end{table}

\noindent\textbf{Design Analysis of CCR.}
Table~\ref{tab:ccr_design} validates the two design decisions of the rectification head made in Sec.~\ref{subsec:coor}. The concatenated input reflects the information structure of offset regression itself, which requires two complementary ingredients: discrete \emph{anchors}, i.e., which cells host separated sources, supplied only by $\tilde{\mathbf{s}}^{(N)}$; and continuous \emph{evidence}, i.e., the sub-cell asymmetry of the diffraction pattern from which the fractional displacement is read, retained only by the pre-sparsification feature $\mathbf{F}_0$. Each feature carries exactly what the other lacks: $\mathbf{F}_0$ leaves overlapping sources unresolved, while $\tilde{\mathbf{s}}^{(N)}$ has erased the sub-cell evidence precisely because sparsification succeeded. The ablation confirms this complementarity: decoding from $\tilde{\mathbf{s}}^{(N)}$ alone, with anchors but no evidence, costs 6.01 CSO-mAP points, 12.50 C-ACC points, and 4.50 dB PSNR. Removing the $\tanh$ bound is subtler: TP-PRMSE is essentially unchanged (0.0457 vs.\ 0.0461) yet CSO-mAP drops; since TP-PRMSE counts only \textit{matched} predictions, the unbounded head must occasionally emit excessive shifts that expel otherwise correct candidates from the matching radius—the survivors remain accurate, but fewer survive. The bounded activation suppresses exactly these outliers, acting as the structural regularizer anticipated in Sec.~\ref{subsec:coor}.

\begin{figure}[htbp]
    \centering
    \includegraphics[width=1.0\columnwidth]{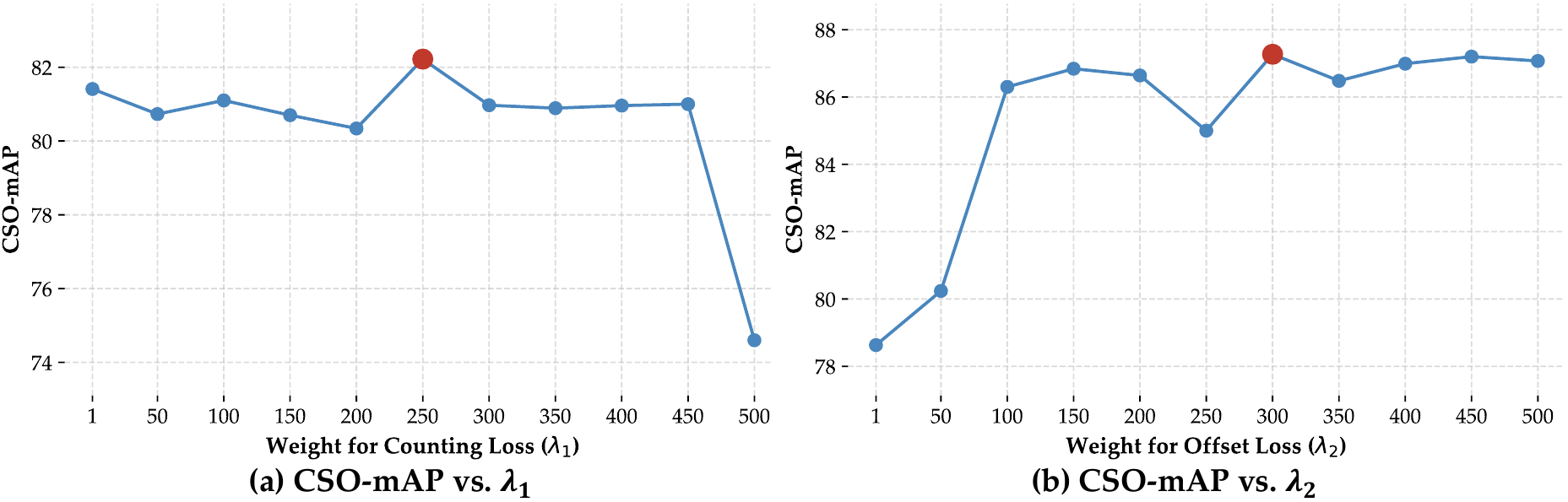}
    \caption{\textbf{Ablation study on loss weight sensitivity.} \textbf{(a) CSO-mAP vs.\ $\lambda_1$.} The effect of the counting loss weight on sub-pixel detection performance. \textbf{(b) CSO-mAP vs.\ $\lambda_2$.} The effect of the offset loss weight. The red dots indicate the optimal settings adopted in our final model ($\lambda_1 = 250$ and $\lambda_2 = 300$). The broad plateaus across both metrics demonstrate that the framework is highly robust and avoids delicate hyperparameter tuning.}
    \label{fig:loss_weights}
\end{figure}

\noindent\textbf{Loss Weight Sensitivity.}
Since the backbone hyperparameters (stage number $N$, branch coefficient $\alpha$, constraint weight $\gamma$) were investigated in our preliminary work~\cite{han2025dista}, we focus on the two new loss weights, sweeping both over $[1, 500]$ (Fig.~\ref{fig:loss_weights}). The counting weight $\lambda_1$ is stable across a wide plateau and peaks at 250; only an extreme setting ($\lambda_1 = 500$) degrades performance, as the count penalty then dominates the gradients of the primary reconstruction objective. The offset weight $\lambda_2$ requires sufficient magnitude, since its mask-driven supervision activates on very few cells and small weights (1 to 50) starve the rectification head of gradient signal; performance rises with $\lambda_2$, plateaus around 300, and remains robust thereafter. We adopt $\lambda_1 = 250$ and $\lambda_2 = 300$ throughout. The breadth of both plateaus indicates that the reported gains do not hinge on delicate tuning.

\section{Conclusion}
\label{sec:conclusion}

This paper traces the stagnation of CSIST unmixing to two defects buried in its inherited formulation: the inference is blind, deprived of any global semantic anchor, and the estimation is discrete, chained to a quantized lattice adopted for tractability rather than fidelity. Our response operates on two levels. At the infrastructure level, we establish the first open research ecosystem for the task, comprising the CSIST-100K benchmark, a decoupled sub-pixel metric suite, and the GrokCSO toolkit. At the algorithmic level, DISTA-Net++ renders unmixing informed, through a count-guided prior that collapses the one-to-many solution space at its source, and continuous, through continuous coordinate rectification that severs localization precision from grid resolution. The experimental evidence matches the diagnosis point for point: the gains concentrate at stringent thresholds where quantization categorically excludes discrete methods, the count prior is most effective where blindness is most costly, and grid densification, the field's traditional route to precision, now degrades rather than improves performance. Sub-pixel precision need no longer be purchased with discretization.

%\section*{Acknowledgments}
%This should be a simple paragraph before the References to thank those individuals and institutions who have supported your work on this article.

%{\appendix[Proof of the Zonklar Equations]
%Use $\backslash${\tt{appendix}} if you have a single appendix:
%Do not use $\backslash${\tt{section}} anymore after $\backslash${\tt{appendix}}, only $\backslash${\tt{section*}}.
%If you have multiple appendixes use $\backslash${\tt{appendices}} then use $\backslash${\tt{section}} to start each appendix.
%You must declare a $\backslash${\tt{section}} before using any $\backslash${\tt{subsection}} or using $\backslash${\tt{label}} ($\backslash${\tt{appendices}} by itself
% starts a section numbered zero.)}


% \input{TPAMI_Main_Document.bbl}
\bibliographystyle{IEEEtran}
\bibliography{TPAMI_Main_Document.bib}

\begin{IEEEbiography}[{\includegraphics[width=1in,height=1.15in,clip,keepaspectratio]{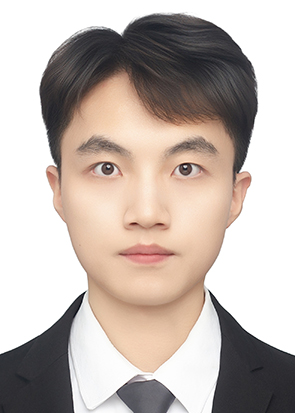}}]
{Mengze Xu}
is currently pursuing the M.E. degree at the College of Computer Science, Nankai University, supervised by Prof. Yimian Dai. He received the B.E. degree from Beijing Foreign Studies University in 2024. His research interests focus on Infrared Small Target Unmixing.
\end{IEEEbiography}

\begin{IEEEbiography}
[{\includegraphics[width=1in,height=1.15in,clip,keepaspectratio]{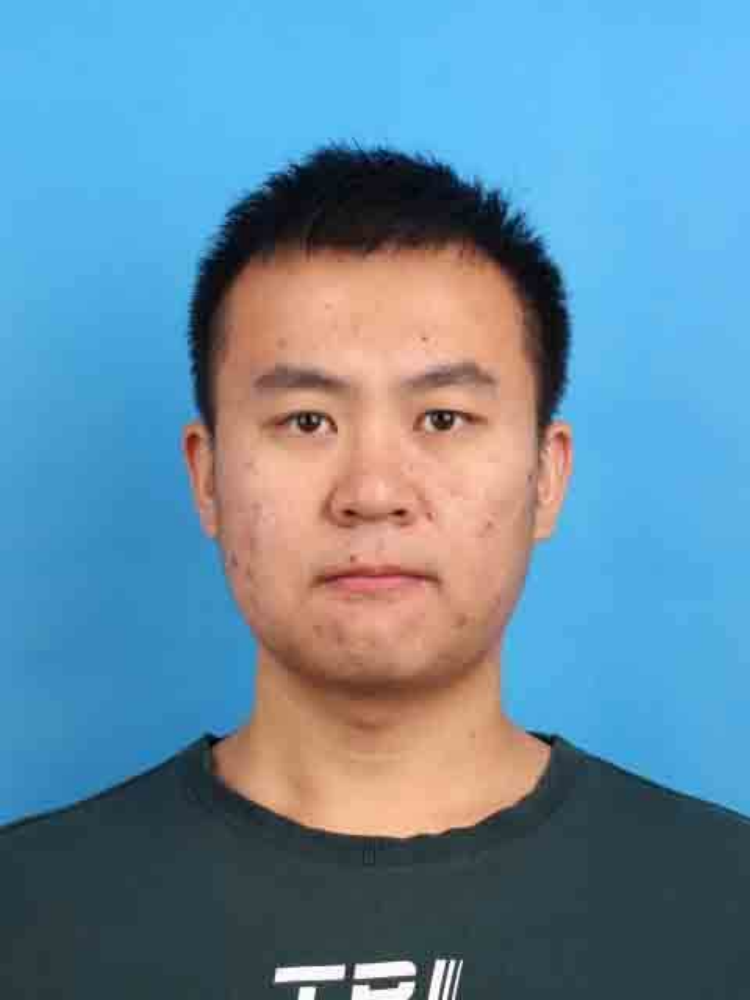}}]
{Zhu Liu} 
    received the B.S. degree in software
	engineering from the Dalian University of Technology, Dalian, China, in 2019. He received his M.S. degree in Software Engineering at Dalian University of Technology, Dalian, China, in 2022. He is  pursuing the Ph.D. degree in Software Engineering at Dalian University of Technology, Dalian, China. His research interests include 
	image processing and optimization.
\end{IEEEbiography}

\begin{IEEEbiography}[{\includegraphics[width=1in,height=1.15in,clip,keepaspectratio]{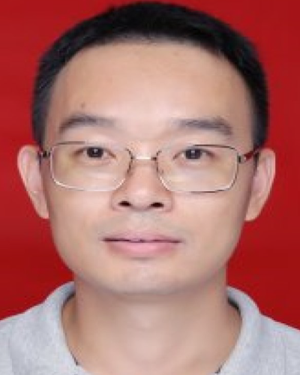}}]
{Weidong Sheng}
received the Ph.D. degree from the National University of Defense Technology (NUDT), Changsha, China, in 2011.
He is currently an Associate Professor with the College of Electronic Science and Technology, NUDT. His current research interests include image processing and data fusion.
\end{IEEEbiography}

\begin{IEEEbiography}[{\includegraphics[width=1in,height=1.15in,clip,keepaspectratio]{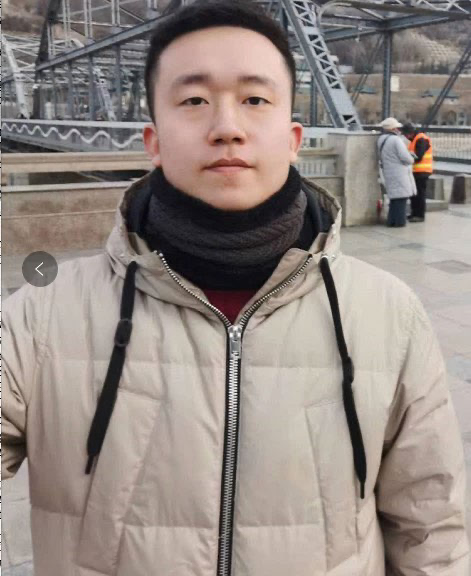}}]
{Boyang Li}
received the B.E. degree in mechanical design manufacture and automation from Tianjin University, Tianjin, China, in 2017, and the M.S. degree in biomedical engineering from the National Innovation Institute of Defense Technology, Academy of Military Sciences, Beijing, China, in 2020. He is currently pursuing the Ph.D. degree in information and communication engineering with the National University of Defense Technology (NUDT), Changsha, China.
His research interests include infrared small target detection, weakly supervised semantic segmentation, and deep learning.
\end{IEEEbiography}

\begin{IEEEbiography}[{\includegraphics[width=1in,height=1.15in,clip,keepaspectratio]{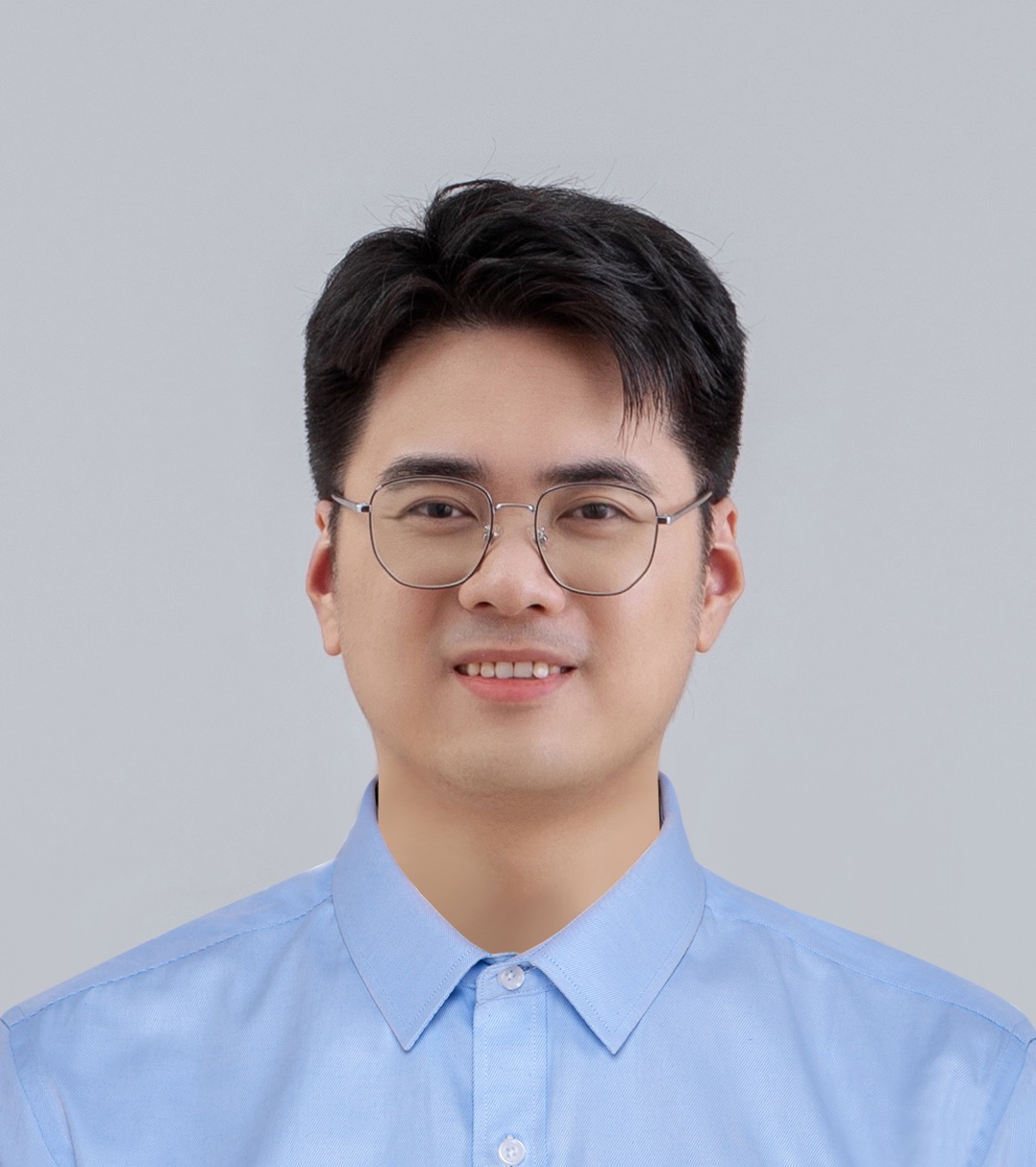}}]
{Yimian Dai}
(Member, IEEE) received the B.E. degree in information engineering and the Ph.D. degree in signal and information processing from Nanjing University of Aeronautics and Astronautics, Nanjing, China, in 2013 and 2020, respectively.
From 2021 to 2024, he was a Postdoctoral Researcher with the School of Computer Science and Engineering, Nanjing University of Science and Technology, Nanjing, China. 
He is currently an Associate Professor with the College of Computer Science, Nankai University, Tianjin, China.
His research interests include computer vision, deep learning, and their applications in remote sensing.
For more information, please visit the link (\href{https://grokcv.site/}{https://grokcv.site/}).
\end{IEEEbiography}

\begin{IEEEbiography}[{\includegraphics[width=1in,height=1.15in,clip,keepaspectratio]{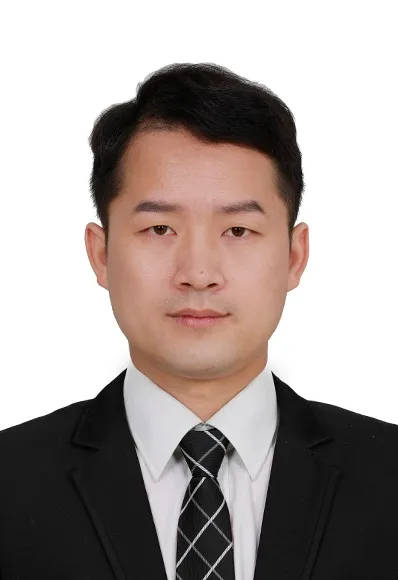}}]
{Ming-Ming Cheng} received his PhD degree from Tsinghua University in 2012.
Then, he did 2 years research fellow, with Prof. Philip Torr in Oxford.
He is now a professor at Nankai University, leading the Media Computing Lab.
His research interests include computer graphics, computer vision, and image processing. 
He received research awards, including the National Science Fund for Distinguished Young Scholars and the ACM China Rising Star Award.
He is on the editorial boards of IEEE TPAMI and IEEE TIP.
\end{IEEEbiography}

\begin{IEEEbiography}[{\includegraphics[width=1in,height=1.15in,clip,keepaspectratio]{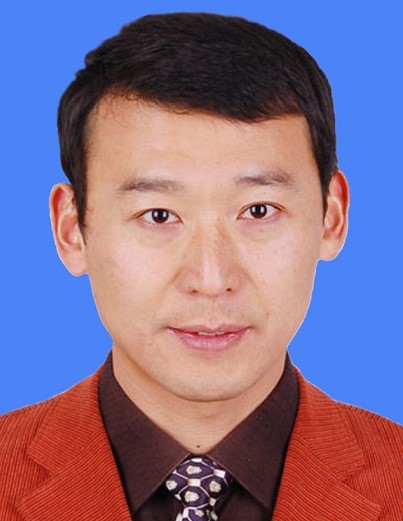}}]
{Jian Yang} received the PhD degree from Nanjing University of Science and Technology (NJUST) in 2002, majoring in pattern recognition and intelligence systems. From 2003 to 2007, he was a Postdoctoral Fellow at the University of Zaragoza, Hong Kong Polytechnic University and New Jersey Institute of Technology, respectively. From 2007 to present, he is a professor in the School of Computer Science and Technology of NJUST. Currently, he is also a visiting distinguished professor in the College of Computer Science of Nankai University. His papers have been cited over 50000 times in the Scholar Google. His research interests include pattern recognition and computer vision. Currently, he is/was an associate editor of Pattern Recognition, Pattern Recognition Letters, IEEE Trans. Neural Networks and Learning Systems, and Neurocomputing. He is a Fellow of IAPR. 
\end{IEEEbiography}

%\newpage

 

% \end{IEEEbiography}



% \vfill

\end{document}